%% file: main.tex
\documentclass[runningheads]{llncs}
\usepackage{graphicx}

\usepackage{tikz}
\usepackage{comment}
\usepackage{amsmath,amssymb} 
\usepackage{color}
\usepackage{wrapfig}
\usepackage{times}
\usepackage{epsfig}
\usepackage{graphicx}
\usepackage{mathtools}
\usepackage{multirow}
\usepackage{bbding}
\usepackage{pifont}
\usepackage{subcaption}

\usepackage{colortbl}

\usepackage[accsupp]{axessibility}  
\usepackage{hyperref}
\hypersetup{pagebackref=true,breaklinks=true,letterpaper=true,colorlinks,bookmarks=false,citecolor=cyan,linkcolor=red}
\newcommand{\Paragraph}[1]{\vspace{0.6mm} \noindent \textbf{#1}
\hspace{0mm}}
\newcommand{\Section}[1]{\vspace{-1mm} \section{#1} \vspace{-1mm}}
\newcommand{\SubSection}[1]{\vspace{-1mm} \subsection{#1} \vspace{-1mm}}

\usepackage{stix}

\newcommand{\Lagr}{\mathcal{L}}
\definecolor{cell_blue}{RGB}{155, 187, 228}

\begin{document}
\pagestyle{headings}
\mainmatter
\def\ECCVSubNumber{7360}  

\title{Multi-domain Learning for Updating Face Anti-spoofing Models} 

\titlerunning{Multi-domain Learning for Updating Face Anti-spoofing Models}
%
\author{Xiao Guo \and
Yaojie Liu \and
Anil Jain \and
Xiaoming Liu}
\authorrunning{X. Guo et al.}
%
\institute{Michigan State University\\
\email{\{guoxia11,liuyaoj1,liuxm,jain\}@cse.msu.edu\\}}
\maketitle

\input{sections/00_abstract}
\input{sections/01_introduction}
\input{sections/02_related_work}
\input{sections/03_01_backbone_study}
\input{sections/03_02_sp_region_estimator}
\input{sections/03_03_FAS_wrapper_architecture}
\input{sections/03_04_training_inference}
\input{sections/04_benchmark}
\input{sections/05_00_exp_setup}

\input{sections/05_01_experiments}
\input{sections/05_02_adaptability}
\input{sections/05_03_exp_sp}
\input{sections/05_04_exp_dual_models}
\input{sections/05_05_cross_domain_exp}
\input{sections/06_conclusion}
\input{sections/07_supplementary}
\clearpage
{\small
\bibliographystyle{splncs04}
\bibliography{egbib}
}
\end{document}

%% file: sections/00_abstract.tex
\begin{abstract}
In this work, we study multi-domain learning for face anti-spoofing (MD-FAS), where a pre-trained FAS model needs to be updated to perform equally well on both source and target domains while only using target domain data for updating.
We present a new model for MD-FAS, which addresses the forgetting issue when learning new domain data, while possessing a high level of adaptability. First, we devise a simple yet effective module, called spoof region estimator (SRE), to identify spoof traces in the spoof image. Such spoof traces reflect the source pre-trained model's responses that help upgraded models combat catastrophic forgetting during updating. Unlike prior works that estimate spoof traces which generate multiple outputs or a low-resolution binary mask, SRE produces one single, detailed pixel-wise estimate in an unsupervised manner. 
Secondly, we propose a novel framework, named FAS-wrapper, which transfers knowledge from the pre-trained models and seamlessly integrates with different FAS models. Lastly, to help the community further advance MD-FAS, we construct a new benchmark based on SIW, SIW-Mv2 and Oulu-NPU, and introduce four distinct protocols for evaluation, where source and target domains are different in terms of spoof type, age, ethnicity, and illumination. Our proposed method achieves superior performance on the MD-FAS benchmark than previous methods. Our code is available at \href{https://github.com/CHELSEA234/Multi-domain-learning-FAS}{https://github.com/CHELSEA234/Multi-domain-learning-FAS}.
\end{abstract}

%% file: sections/01_introduction.tex
\section{Introduction}
Face anti-spoofing (FAS) comprises techniques that distinguish genuine human faces and faces on spoof mediums~\cite{face-anti-spoofing-using-patch-and-depth-based-cnns}, such as printed photographs, screen replay, and 3D masks. 
FAS is a critical component of the face recognition pipeline that ensures only genuine faces are being matched.
As face recognition systems are widely deployed in real world applications, a laboratory-trained FAS model is often required to deploy in a new target domain with face images from novel camera sensors, ethnicities, ages, types of spoof attacks, {\it etc.}, which differ from the source domain training data in the laboratory.

In the presence of a large domain-shift \cite{saenko2010adapting,torralba2011unbiased,khosla2012undoing} between the source and target domain, it is necessary to employ new target domain data for updating the pre-trained FAS model, in order to perform well in the new test environment. Meanwhile, the source domain data might be inaccessible during updating, due to data privacy issues, which happens more and more frequently for Personally Identifiable Information (PII). Secondly, the FAS model needs to be evaluated jointly on source and target domains, as spoof attacks should be detected regardless of which domain they originate from. Motivated by these challenges, the goal of this paper is to answer the following question:

\begin{center}
\textit{How can we update a FAS model using only target domain data, so that the upgraded model can perform well in both the source and target domains?}
\end{center}


\input{sections/figure_table_latex/figure_1_overview}

We define this problem as multi-domain learning face anti-spoofing (MD-FAS), as depicted in Fig.~\ref{fig:pro_definition}. 
Notably, Domain Adaptation (DA) works \cite{ganin2016domain,saito2018open,liu2019separate,kundu2020class,long2015learning} mainly evaluate on the target domain, whereas MD-FAS requires a joint evaluation. Also, MD-FAS is related to Multiple Domain Learning (MDL) \cite{rebuffi2017learning,rebuffi2018efficient,guo2019depthwise}, which aims to learn a universal representation for images in many generic image domains, based on one unchanged model. In contrast, MD-FAS algorithm needs to be model-agnostic for the deployment, which means the MD-FAS algorithm can be tasked to update FAS models with various architectures or loss functions. Lastly, the source domain data is unavailable during the training in MD-FAS, which is different from previous domain generalization methods in FAS~\cite{shao2019multi,liu2019deep,jia2020single,qin2020learning} or related manipulation detection problems~\cite{proactive-image-manipulation-detection}.

There are two main challenges in MD-FAS. First, the source domain data is unavailable during the updating. As a result, MD-FAS easily suffers from the long-standing \textit{catastrophic forgetting}~\cite{kirkpatrick2017overcoming} in learning new tasks, gradually degrading source domain performance. The most common solution \cite{li2017learning,dhar2019learning,jung2016less} to such a forgetting issue is to use logits and class activation map (grad-CAM)~\cite{selvaraju2017grad} restoring prior model responses when processing the new data. However, due to the increasingly sophisticated spoof image, using logits and grad-CAM empirically fail to precisely pinpoint spatial pixel locations where spoofness occurs, unable to uncover the decision making behind the FAS model. To this end, we propose a simple yet effective module, namely \textit{spoof region estimator} (\textit{SRE}), to identify the spoof regions given an input spoof image. Such spoof traces serve as responses of the pre-trained model, or better replacement to logits and activation maps in the MD-FAS scenario. Notably, unlike using multiple traces to pinpoint spoofness or manipulation in image~\cite{liu2019separate,zhao2021multi}, or low-resolution binary mask as manipulation indicator~\cite{on-the-detection-of-digital-face-manipulation,yu2021revisiting,pssc-net-progressive-spatio-channel-correlation-network-for-image-manipulation-detection-and-localization}, our \textit{SRE} offers a single and high-resolution detailed binary mask representing pixel-wise spatial locations of spoofness. 
Also, many anti-forgetting algorithms~\cite{rebuffi2017icarl,chaudhry2018efficient,shin2017continual,rusu2016progressive,mallya2018packnet,fernando2017pathnet} usually require extra memory for restoring exemplar samples or expanding the model size, which makes them inefficient in real-world situations.

Secondly, to develop an algorithm with a high level of adaptability, it is desirable to keep original FAS models intact for the seamless deployment while changing the network parameters. Unlike methods proposed in \cite{rebuffi2017learning,rebuffi2018efficient} that specialize on the certain architecture (\textit{e.g.}, ResNet), we first derive the general formulation after studying FAS models~\cite{yu2020face,yu2020searching,liu2018learning,liu2020disentangling,liu2021casia,shao2019multi}, then based on such a formulation we propose a novel architecture, named \textit{FAS-wrapper} (depicted in Fig.~\ref{fig:overall_archi}), which can be deployed for FAS models with minimum changes on the architecture.

In summary, this paper makes the following contributions:

$\diamond$ Driven by the deployment in real-world applications, we define a new problem of MD-FAS, which requires to update a pre-trained FAS model only using target domain data, yet evaluate on both source and target domains. 
To facilitate the MD-FAS study, we construct the FASMD benchmark, based on existing FAS datasets~\cite{liu2019deep,liu2018learning,OULU_NPU_2017}, with four evaluation protocols. 

$\diamond$ We propose a \textit{spoof region estimator} (\textit{SRE}) module to identify spoof traces in the input image. Such spoof traces serve as the prior model's responses to help tackle the \textit{catastrophic forgetting} during the FAS model updating.

$\diamond$ We propose a novel method, \textit{FAS-wrapper}, which can be adopted by any FAS models for adapting to target domains while preserving the source domain performance.

$\diamond$ Our method demonstrates superior performance over prior works, on both source and target domains in the FASMD benchmark.
Moreover, our method also generalizes well in the cross-dataset scenario.

%% file: sections/figure_table_latex/figure_1_overview.tex
\begin{figure}[t]
 \centering
 \includegraphics[height=4cm,width=11cm]{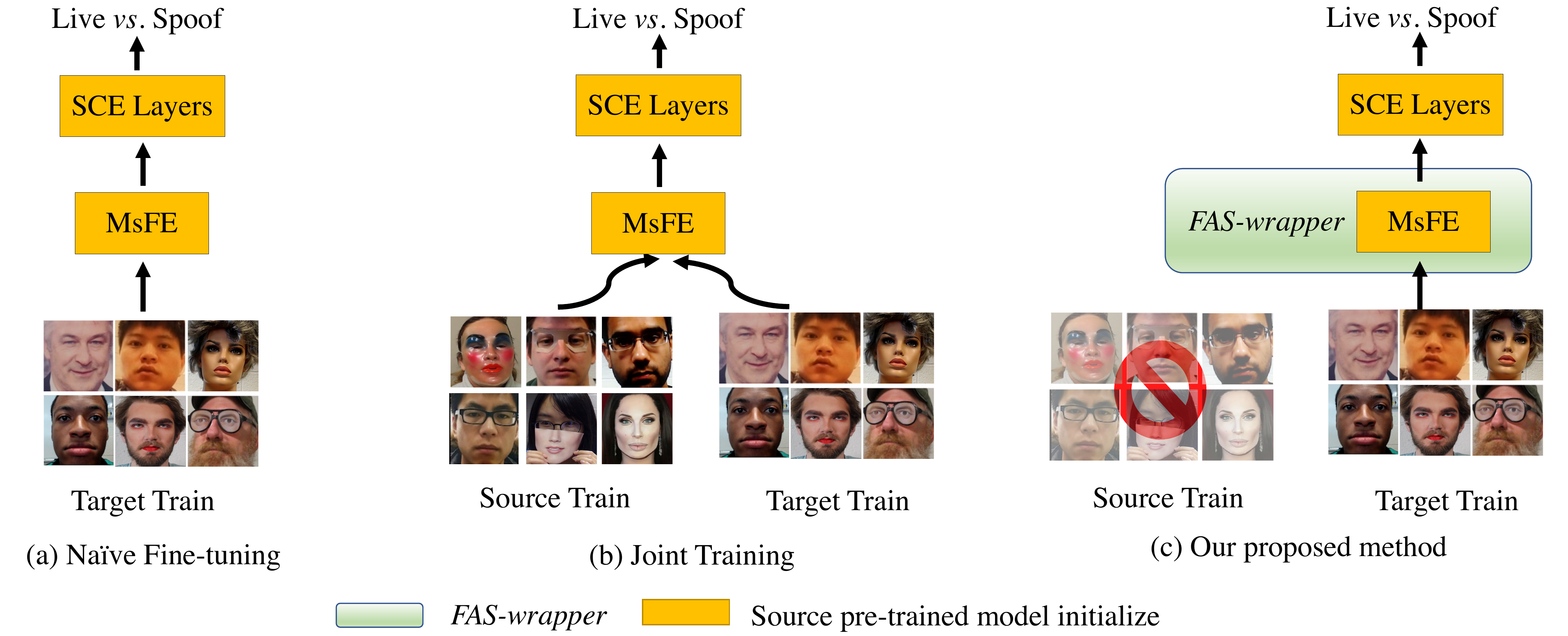}
 \vspace{-3mm}
 \caption{\footnotesize We study multi-domain learning face anti-spoofing (MD-FAS), in which the model is trained only using target domain data. We first derive the general formulation of FAS models, which contains Spoof Cue Estimate Layers (SCE layers) and multi-scale feature extractor (MsFE). Based on these two components, we propose \textit{FAS-wrapper} that can be adopted for any FAS models, as depicted in Fig.~1(c). Fig.~1(a) and 1(b) represent the naive fine-tuning and joint training. \vspace{-3mm}}
 \label{fig:pro_definition}
\end{figure}

%% file: sections/02_related_work.tex
\input{sections/figure_table_latex/table_1_comparison}

\Section{Related Works}
\Paragraph{Face Anti-spoofing Domain Adaptation} In Domain Adaption (DA)~\cite{ganin2016domain,saito2018open,liu2019separate,kundu2020class,long2015learning}, many prior works assume the source data is accessible, but in our setup, source domain data is unavailable. The DA performance evaluation is biased towards the target domain data, as source domain performance may deteriorate, whereas FAS models need to excel on both source and target domain data.
There are some FAS works that study the cross-domain scenario~\cite{yang2021few,shao2019multi,jia2020single,liu2019deep,quan2021progressive,tu2020learning,noise-modeling-synthesis-and-classification-for-generic-object-anti-spoofing}. \cite{yang2021few} is proposed for the scenario where source and a few labeled new domain data are available, with the idea to augment target data by style transfer~\cite{yoo2019photorealistic}. \cite{shao2019multi} learns a shared, indiscriminative feature space without the target domain data. Besides, \cite{jia2020single} constructs a generalized feature space that has a compact real faces feature distribution in different domains. \cite{liu2019deep} also works on unseen domain generalization. But the same as the other works, the new domain is not based on bio-metric patterns (\textit{i.e.}, age). 
Being orthogonal to prior works, the source domain data in our study is unavailable, which is a more challenging setting, as shown in Tab.~\ref{tab_background_compare}.

\input{sections/figure_table_latex/figure_2_overall_archi}

\Paragraph{Anti-forgetting Learning}
\label{background_study_spoof}
The main challenge in MD-FAS is the long-studied \textit{catastrophic forgetting} \cite{kirkpatrick2017overcoming}. According to ~\cite{delange2021continual},
there exist four solutions:
replay~\cite{rebuffi2017icarl,chaudhry2018efficient,shin2017continual}, parameter isolation~\cite{rusu2016progressive,mallya2018packnet,fernando2017pathnet}, prior-driven~\cite{kirkpatrick2017overcoming,lee2017overcoming,aljundi2018memory} and data-driven~\cite{li2017learning,dhar2019learning,jung2016less}. The replay method requires to restore a fraction of training data which breaks our source-free constraint, 
\textit{e.g.}, \cite{rostami2021detection} 
needs to store the exemplar training data. Parameter isolation methods~\cite{rusu2016progressive,mallya2018packnet,fernando2017pathnet} dynamically expand the network, which is also discouraged due to the memory expense. 
The prior-driven methods~\cite{kirkpatrick2017overcoming,lee2017overcoming,aljundi2018memory} are proposed 
based on the assumption that model parameters obey the Gaussian distribution, which is not always the case. 
The data-driven method~\cite{guo2019human,guo2020cord19sts,hsu2021discourse,abdalmageed2020assessment} is always more favored in the community, due to its effectiveness and low computation cost. However, the development of data-driven methods is dampened in the FAS, since the commonly-used pre-trained model responses (\textit{e.g.}, class probabilities~\cite{li2017learning} and grad-CAM~\cite{dhar2019learning}) fail to capture spoof regions.
In this context, our SRE is a simple yet effect way of estimating the spoof trace in the image, which serves as the responses of the pre-train model.

\Paragraph{Multi-domain Learning}
\label{background_study_domain}
Mostly recently, many large-scale FAS datasets with rich annotations have been collected~\cite{zhang2020celeba,zhang2020casia,liu2021casia} in the community, among which \cite{liu2021casia} studies cross-ethnicity and cross-gender FAS. 
However they work on multi-modal datasets, whereas our input is a single RGB image.
In the literature, our work is similar to the multi-domain learning (MDL) \cite{rebuffi2017learning,rebuffi2018efficient,mancini2018adding}, where a re-trained model is required to perform well on both source and target domain data. The common approaches are proposed from~\cite{rebuffi2017learning,rebuffi2018efficient} based on ResNet~\cite{he2016deep}, which, compared to \cite{simonyan2014very,krizhevsky2012imagenet}, has advantages in increasing the abstraction by convoluation operations. 
In contrast, 
an ideal MD-FAS algorithm, such as \textit{FAS-wrapper}, should work in a  model-agnostic fashion.

%% file: sections/figure_table_latex/table_1_comparison.tex
\begin{table*}[t!]
\centering
\scriptsize
\resizebox{1\textwidth}{!}{
\begin{tabular}{c|c|c|c|c|c|c}
\hline
 Paradigm & Method & Source Free & \begin{tabular}[c]{@{}c@{}}Learning \\New Domain\end{tabular} & \begin{tabular}[c]{@{}c@{}}Joint\\Evaluation\end{tabular} & \begin{tabular}[c]{@{}c@{}}Model \\Agnostic\end{tabular}& \begin{tabular}[c]{@{}c@{}}Anti-forgetting\\ Mechanism\end{tabular}\\ \hline

\multirow{3}{*}{\begin{tabular}[c]{@{}c@{}}Face Anti-Spoofing\\ Domain Learning\end{tabular}} 
& \begin{tabular}[c]{@{}l@{}} SSDG \cite{jia2020single}\end{tabular} & \textcolor{red}{\ding{56}} & \textcolor{green}{\ding{52}} & \textcolor{green}{\ding{52}} & \textcolor{green}{\ding{52}} & N/A\\ \cline{2-7} 
& \begin{tabular}[c]{@{}l@{}} MADDoG \cite{shao2019multi}\end{tabular} & \textcolor{red}{\ding{56}} & \textcolor{green}{\ding{52}} & \textcolor{green}{\ding{52}} & \textcolor{green}{\ding{52}} & N/A\\ \cline{2-7} 
& \begin{tabular}[c]{@{}l@{}} FSDE-FAS \cite{yang2021few}\end{tabular} & \textcolor{red}{\ding{56}} & \textcolor{green}{\ding{52}} & \textcolor{green}{\ding{52}} & \textcolor{green}{\ding{52}} & N/A\\ 
\hline

\multirow{5}{*}{\begin{tabular}[c]{@{}c@{}}Anti-forgetting\\Learning\end{tabular}} 
& EWC~\cite{kirkpatrick2017overcoming} & \textcolor{green}{\ding{52}} & \textcolor{red}{\ding{56}} & \textcolor{green}{\ding{52}} & \textcolor{green}{\ding{52}} & Prior-driven\\ \cline{2-7} 
 & iCaRL~\cite{rebuffi2017icarl} & \textcolor{green}{\ding{52}} & \textcolor{red}{\ding{56}} & \textcolor{green}{\ding{52}} & \textcolor{green}{\ding{52}} & Replay\\ \cline{2-7} 
 & MAS~\cite{aljundi2018memory} & \textcolor{green}{\ding{52}} & \textcolor{red}{\ding{56}} & \textcolor{green}{\ding{52}} & \textcolor{green}{\ding{52}} & Prior-driven\\ \cline{2-7} 
 & LwF~\cite{li2017learning} & \textcolor{green}{\ding{52}} & \textcolor{red}{\ding{56}} & \textcolor{green}{\ding{52}} & \textcolor{green}{\ding{52}} & Data-driven (class prob.)\\ \cline{2-7}  
 & LwM~\cite{dhar2019learning} & \textcolor{green}{\ding{52}} & \textcolor{red}{\ding{56}} & \textcolor{green}{\ding{52}} & \textcolor{green}{\ding{52}} & Data-driven (feat. map) \\ \hline
 
\multirow{6}{*}{\begin{tabular}[c]{@{}c@{}}Multi-domain\\Learning\end{tabular}} 
 & DAN~\cite{ganin2016domain} & \textcolor{red}{\ding{56}} & \textcolor{green}{\ding{52}} & \textcolor{red}{\ding{56}} & \textcolor{green}{\ding{52}} & N/A\\ \cline{2-7} 
 & OSBP~\cite{saito2018open} & \textcolor{red}{\ding{56}} & \textcolor{green}{\ding{52}} & \textcolor{red}{\ding{56}} & \textcolor{green}{\ding{52}} & N/A\\ \cline{2-7} 
 & STA~\cite{liu2019separate} & \textcolor{red}{\ding{56}} & \textcolor{green}{\ding{52}} & \textcolor{red}{\ding{56}} & \textcolor{green}{\ding{52}} & N/A\\ \cline{2-7} 
 & CIDA~\cite{kundu2020class} & \textcolor{green}{\ding{52}} & \textcolor{green}{\ding{52}} & \textcolor{red}{\ding{56}} & \textcolor{red}{\ding{56}} & N/A\\ \cline{2-7} 
 & \begin{tabular}[c]{@{}l@{}}Seri. Adapter~\cite{rebuffi2017learning}\end{tabular} & \textcolor{green}{\ding{52}} & \textcolor{green}{\ding{52}} & \textcolor{green}{\ding{52}} & \textcolor{red}{\ding{56}} & N/A\\ \cline{2-7} 
 &  \begin{tabular}[c]{@{}l@{}}Para. Adapter~\cite{rebuffi2018efficient}\end{tabular} & \textcolor{green}{\ding{52}} & \textcolor{green}{\ding{52}} & \textcolor{green}{\ding{52}} & \textcolor{red}{\ding{56}} & N/A\\ \hline
 
\begin{tabular}[c]{@{}c@{}}Multi-domain Learning \\ Face Anti-spoofing  \end{tabular} & \textit{FAS-wrapper} (Ours) & \textcolor{green}{\ding{52}} & \textcolor{green}{\ding{52}} & \textcolor{green}{\ding{52}} & \textcolor{green}{\ding{52}} & Data-driven (spoof region) \\ \hline
\end{tabular}}
\vspace{1mm}
\caption{\footnotesize We study the multi-domain learning face anti-spoofing, which is different to prior works. \vspace{-7mm}}
\label{tab_background_compare}
\end{table*}

%% file: sections/figure_table_latex/figure_2_overall_archi.tex
\begin{figure*}[t]
 \centering
\includegraphics[height=5cm]{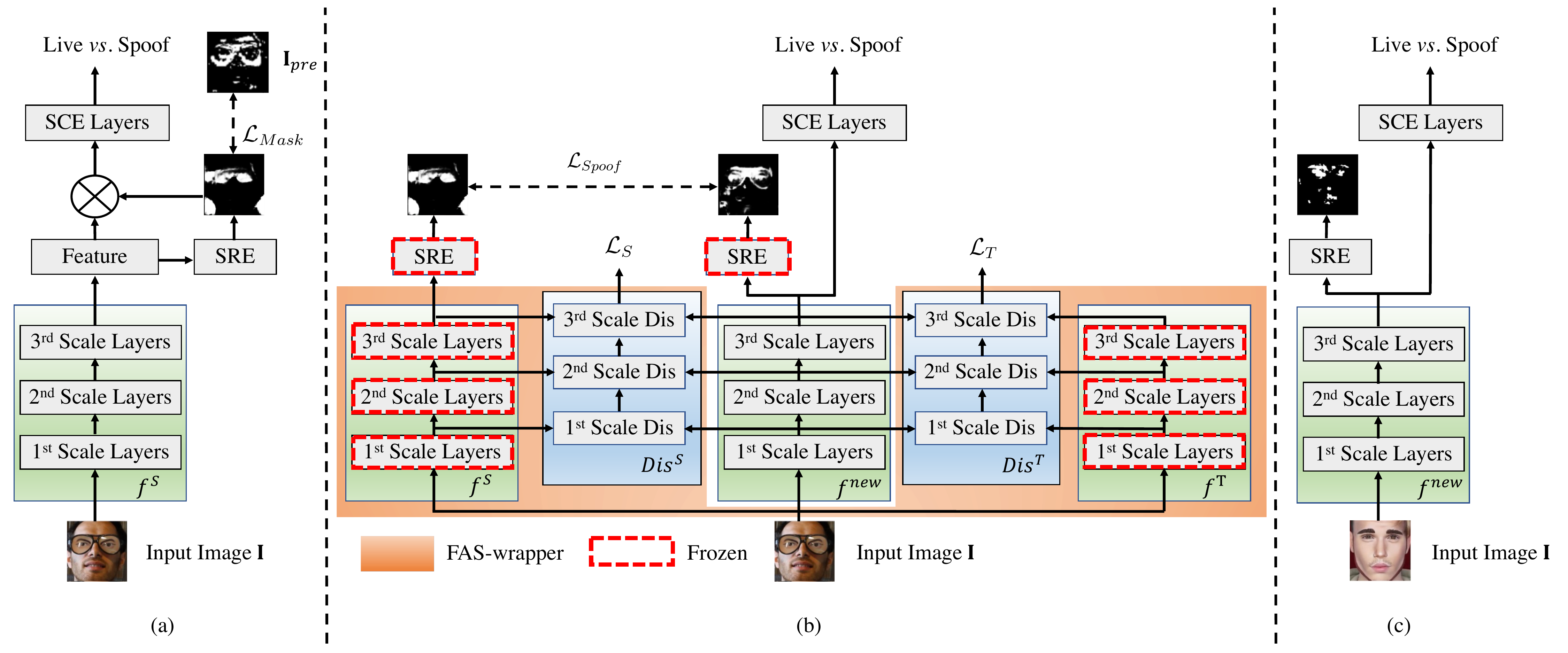}
\vspace{-3mm}
 \caption{\footnotesize (a) Given the source pre-trained model that contains feature extractor $f^{S}$, we fine-tune it with the proposed \textit{spoof region estimator} (SRE) on the target domain data, in which we use preliminary mask ($\textbf{I}_{pre}$) to assist the learning (see Sec.~\ref{sec_sp_est}). Then, we obtain a well-trained \textit{SRE} and a new feature extractor $f^{T}$ which specializes in the target domain. (b) In \textit{FAS-wrapper}, SRE helps $f^{S}$ and updated model ($f^{new}$) generate binary masks indicating spoof cues, which serve as model responses given an input image ($\textbf{I}$). $\Lagr_{Spoof}$ prevents the divergence between estimated spoof traces, to combat \textit{catastrophic forgetting}. Meanwhile, using two multi-scale discriminators ($Dis^{S}$ and $Dis^{T}$), \textit{FAS-wrapper} transfers the knowledge from two teacher models ( $f^{S}$ and $f^{T}$) to $f^{new}$ via the adversarial training. (c) The update model $f^{new}$ and SRE can be used for the inference.\vspace{-3mm}}
 \label{fig:overall_archi}
\end{figure*}


%% file: sections/03_01_backbone_study.tex
\Section{Proposed Method}
\label{sec_architecture}
This section is organized as follows. Sec.~\ref{sec_background_study_backbone} summarizes the general formulation of recent FAS models.  Sec.~\ref{sec_sp_est} and \ref{sec_multi_dis} introduce the \textit{spoof region estimator} and overall \textit{FAS-wrapper} architecture. Training and inference procedures are reported in Sec.~\ref{sec_training_and_inference}.

\SubSection{FAS Models Study}
\label{sec_background_study_backbone}
We investigate the recently proposed FAS methods (see Tab.~\ref{tab_FAS_compare}) and observe that these FAS models have two shared characteristics.

\Paragraph{Spoof Cue Estimate} Beyond treating FAS as a binary classification problem, many SOTA works emphasize on estimating spoof clues from a given image. 
Such spoof clues are detected in two ways: (a) optimizing the model to predict auxiliary signals such as depth map or rPPG signals~\cite{liu2018learning,yu2020searching,yu2020face}; (b) interpreting the spoofness from different perspectives:
the method in~\cite{jourabloo2018face} aims to disentangle the spoof noise, including color distortions and different types of artifacts, and spoof traces are interpreted in~\cite{liu2020physics,liu2020disentangling} as multi-scale and physical-based traces.

\input{sections/figure_table_latex/table_2_backbone_study}
\Paragraph{Multi-scale Feature Extractor} Majority of previous FAS methods adopt the multi-scale feature. 
We believe such a multi-scale structure assists in learning information at different frequency levels. 
This is also demonstrated in~\cite{liu2020disentangling} that low-frequency traces (\textit{e.g.}, makeup strokes and specular highlights) and high-frequency content (\textit{e.g.}, Moiré patterns) are equally important for the FAS models' success. 

As a result, we formalize the generic FAS model using two components: feature extractor $f$ and spoof cue estimate (SCE) layers (or decoders) $g$. 
When $f$ takes an input face image, denoted as $\mathbf{I}$, the output feature map at $t$-th layer of the feature extractor $f$ is $f_{t}(\mathbf{I})$. The size of $f_{t}(\mathbf{I})$ is $C_{t} \times H_{t} \times W_{t}$, where $C_{t}$ is the channel number, and $H_{t}$ and $W_{t}$ are respectively the height and width of feature maps. 

%% file: sections/figure_table_latex/table_2_backbone_study.tex
\begin{wraptable}{r}{0.6\textwidth} 
\vspace{-4mm}
\centering
\scriptsize
\begin{tabular}{c|c|c|c}
\hline
\textbf{Method}&\textbf{Year}&\textbf{Number of Scale}&\textbf{Spoof Cue Estimate} \\ \hline
Auxiliary \cite{liu2018learning} & $2018$ & $3$ & Depth and rPPG signal \\ \hline
Despoofing \cite{jourabloo2018face} & $2018$ & $3$ & \begin{tabular}[c]{@{}c@{}}Color distortions, \\ and display artifacts\end{tabular} \\ \hline
MADD \cite{shao2019multi} & $2019$ & $3$ & Depth \\ \hline
CDCN \cite{yu2020searching} & $2020$ & $3$ & Depth \\ \hline
STDN \cite{liu2020disentangling} & $2020$ & $3$ & \begin{tabular}[c]{@{}c@{}}Color range bias, content and\\ texture pattern, and depth.\end{tabular} \\ \hline
BCN \cite{yu2020face} & $2020$ & $3$ & \begin{tabular}[c]{@{}c@{}}Patch, reflection and depth.\end{tabular} \\ \hline
PSMM-Net \cite{liu2021casia} & $2021$ & $4$ & \begin{tabular}[c]{@{}c@{}}Depth, RGB and infrared image.\end{tabular} \\ \hline
PhySTD \cite{liu2020physics} & $2022$ & $4$ & \begin{tabular}[c]{@{}c@{}}Additive and inpainting trace,\\ and depth.\end{tabular} \\ \hline
\end{tabular}
\caption{\footnotesize Summary of recent FAS models.
\vspace{-36mm}}
\label{tab_FAS_compare}
\end{wraptable} 

%% file: sections/03_02_sp_region_estimator.tex
\SubSection{Spoof Region Estimator}
\label{sec_sp_est}
\vspace{-0.1cm}
\Paragraph{Motivation}
Apart from the importance of identifying spoof cues for FAS performance, we observe that spoof trace also serves as a key reflection of how different models make the binary decision, namely, different models' activations on the input image. 
In other words, although different models might unanimously classify the same image as spoof, they in fact could make decisions based on distinct spatial regions, as depicted in Fig.~\ref{fig_reflection}. 
Thus, we attempt to prevent the divergence between spoof regions estimated from the new model (\textit{i.e.}, $f^{new}$) and source domain pre-trained model (\textit{i.e.}, $f^{S}$), such that we can enable $f^{new}$ to perceive spoof cues from the perspective of $f^{S}$, thereby combating the \textit{catastrophic forgetting} issue. To this end, we propose a \textit{spoof region estimator} (\textit{SRE}) to localize spatial pixel positions with spoof artifacts or covered by spoof materials.

\input{sections/figure_table_latex/figure_3_preliminary_mask}

\Paragraph{Formulation} Let us formulate the spoof region estimate task. 
We denote the pixel collection in an image as $D_{\textbf{I}} = \{{(x_1,y_1),(x_2,y_2),...,(x_n,y_n)}\}$, the proposed method aims to predict the region where the area of presentation attack can be represented as a binary mask, denoted as $D_{\textit{pred}} = \{{(x_1,y'_1),(x_2,y'_2),...,(x_n,y'_n)}\}$, where $x_{i}$, $y_{i}$ and $y'_{i}$ respectively represent the pixel, ground truth pixel label, and predicted label at $i$~th pixel. 
Also, the spoof region estimate task can be regarded as a pixel-level binary classification problem, namely pixel being live or spoof, thus we have $y_{i} \in \{ \textit{o}^{Live}, \textit{o}^{Spoof}\}$.
Note that $i \in \{ 1,2,3...,n \}$ and $n$ is the total number of pixels in the image. 

\Paragraph{Method} 
As depicted in Fig.~\ref{fig:overall_archi}, we insert a \textit{SRE} module in the source pre-trained model, between the feature extractor $f^{S}$ and spoof cue estimate layers $g^{S}$. The region estimator converts $f^{s}(\textbf{I})$ to a binary mask $\textbf{M}$ with the size $H_{t'} \times W_{t'}$.
In the beginning of the training, we create the preliminary mask to supervise \textit{SRE} for generating the spoof region. 
The preliminary mask generation is based on the reconstruction method proposed in~\cite{liu2020disentangling}, as illustrated in Fig.~\ref{fig_preliminary_mask_process}. In particular, we denote input spoof image as $\textbf{I}_{spoof}$ and use the method in \cite{liu2020disentangling} to reconstruct its live counterpart $\hat{I}_{live}$. By subtracting $\textbf{I}_{spoof}$ from $\hat{\textbf{I}}_{live}$, and taking the absolute value of the resulting image, we obtain the different image $\textbf{I}_{d}$,
whose size is $C_{0} \times H_{0} \times W_{0}$ where $C_{0}$ is $3$. 
We convert $\textbf{I}_{d}$ to a gray image $\hat{\textbf{I}}_{d}$, by summing along with its channel dimension. Apparently, $\hat{\textbf{I}}_{d}$ has the size as $C_{1} \times H_{0} \times W_{0}$ where $C_{1}$ is 1. 
We assign each pixel value in the preliminary mask by applying a predefined threshold $T$, 
\begin{equation}
\footnotesize
  p'_{ij} =
  \begin{cases}
   0 & p_{ij} < T \\
   1 & p_{ij} \ge T, \\
  \end{cases}
\end{equation}
where pixels in $\hat{\textbf{I}}_{d}$ and $\textbf{I}_{pre}$ are $p_{ij}$ and $p'_{ij}$ respectively.

Evidently, the supervisory signal $\textbf{I}_{pre}$ is not the ground truth. 
Inspired by~\cite{on-the-detection-of-digital-face-manipulation} that a model can generate the manipulation mask by itself during training procedure,
we only use $\textbf{I}_{pre}$ as the supervision at the first a few training epochs, then steer the model itself to find the optimal spoof region by optimizing towards a higher classification accuracy.  
More details are in Sec.~\ref{sec_training_and_inference}.

\Paragraph{Discussion}
Firstly, we discuss the difference to prior spoof region estimate works. The previous methods~\cite{liu2020disentangling,liu2020physics} use various traces to help live or spoof image reconstruction, while our goal is to pinpoint the region with spoof artifacts, which serves as pre-trained model's responses to help the new model behave similar to the pre-trained one$($s$)$, alleviating the forgetting issue.
\cite{on-the-detection-of-digital-face-manipulation} offers low-resolution binary masks as the supervisory signal, but our self-generated $\textbf{I}_{pre}$ can only bootstrap the system.
Also, \cite{zhao2021multi} proposes an architecture for producing multiple masks, which is not practical in our scenario. 
Thus our mask generation method is different from theirs.
Finally, SRE can be a plug-in module for any given FAS model, and details are in Sec.~\ref{sec_exp_sp}.

%% file: sections/figure_table_latex/figure_3_preliminary_mask.tex
\begin{figure}[t]
 \centering
 \includegraphics[scale=0.4]{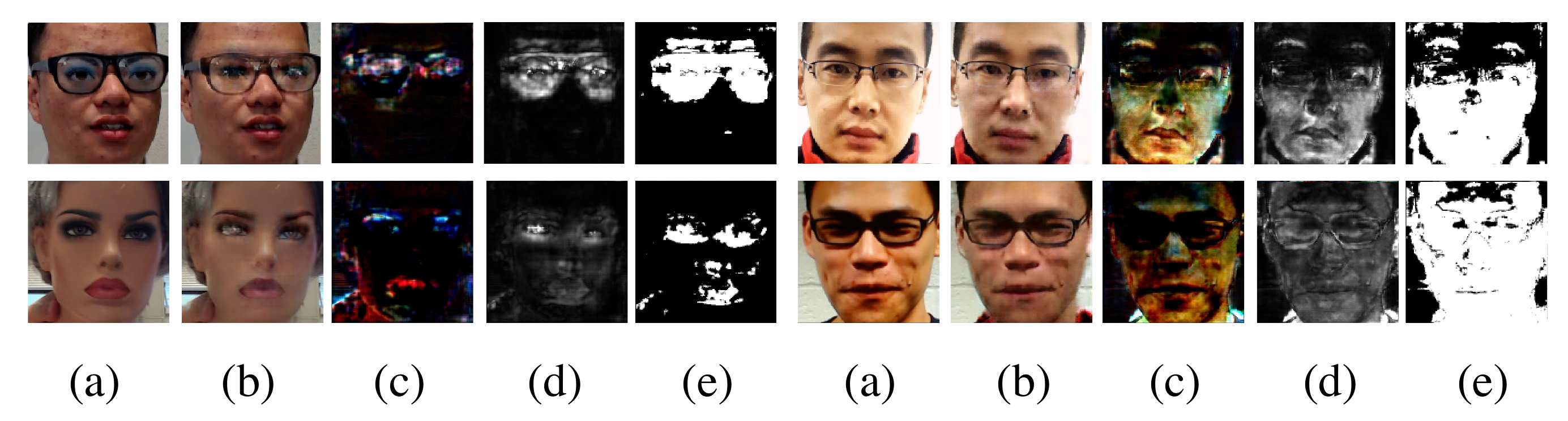}
 \label{fig_preliminary_mask}
 \vspace{-3mm}
 \caption{\footnotesize The preliminary mask generation process: (a) the spoof image, (b) the live reconstruction, (c) and (d) are difference image in RGB and gray format, and (e) is the preliminary spoof mask.\vspace{-4mm}}
\label{fig_preliminary_mask_process}
\end{figure}

%% file: sections/03_03_FAS_wrapper_architecture.tex
\SubSection{\textit{FAS-wrapper} Architecture}
\label{sec_multi_dis}
\Paragraph{Motivation} We aim to deliver an update algorithm that can be effortlessly deployed to different FAS models. Thus, it is important to design a model agnostic algorithm that allows the FAS model to remain intact, thereby maintaining the original FAS model performance. 
Our \textit{FAS-wrapper} operates in a model-agnostic way where only external expansions are made, largely maintaining the original FAS model's ability.

As depicted in Fig.~\ref{fig:overall_archi}, we denote the source pre-trained feature extractor $f^{S}$ as \textit{source teacher}, and the feature extractor after the fine-tuning procedure as \textit{target teacher} ($f^{T}$). Instead of using one single teacher model like~\cite{jia2020single}, we use $f^{S}$ and $f^{T}$ to regularize the training, offering the more informative and instructive supervision for the newly upgraded model, denoted as $f^{new}$.
Lastly, unlike prior FAS works~\cite{jia2020single,shao2019multi,yang2021few} which apply the indiscriminative loss on the final output embedding or logits from $f^{S}$, we construct multi-scale discriminators that operate at the feature-map level for aligning intermediate feature distributions of $f^{new}$ to those of teacher models (\textit{i.e.}, $f^{T}$ and $f^{S}$). Motivations of the multi-scale discriminators are: (a) the multi-scale features, as a common FAS model attribute (Sec.~\ref{sec_background_study_backbone}), should be considered; (b) the adversarial learning can be used at the feature-map level which contains the richer information than final output logits. 

\Paragraph{Method} We construct two multi-scale discriminators, $Dis^{S}$ and $Dis^{T}$, for transferring semantic knowledge from $f^{S}$ and $f^{T}$ to $f^{new}$ respectively, via an adversarial learning loss. Specifically, at $l$-th scale, $Dis_{l}^{S}$ and $Dis_{l}^{T}$ take the previous discriminator output and the $l$-th scale feature generated from feature extractors. 
We use ${\mathbf{d}}^{S}_{l}$ and ${\mathbf{d}}^{T}_{l}$ to represent two discriminators' outputs at $l$-th level while taking teacher generated features (\textit{i.e.}, $f^{S}_{l}(\textbf{I})$ and $f^{T}_{l}(\textbf{I})$), and $\mathbf{d}^{\prime S}_{l}$ and $\mathbf{d}^{\prime T}_{l}$ while taking upgraded model generated feature, $f^{new}_{l}(\textbf{I})$. Therefore, the first-level discriminator output are:
\begin{equation}
    \footnotesize
    \mathbf{d}^{S}_{1} = Dis_{1}^{S}(f^{S}_{1}(\textbf{I})), \quad
    \textbf{d}^{T}_{1} = Dis_{1}^{T}(f^{T}_{1}(\textbf{I})),
\end{equation}
\begin{equation}
    \footnotesize
    \textbf{d}^{\prime S}_{1} = Dis_{1}^{S}(f^{new}_{1}(\textbf{I})),
    \quad 
    \textbf{d}^{\prime T}_{1} = Dis_{1}^{T}(f^{new}_{1}(\textbf{I})),
\end{equation}
and discriminators at following levels take the $l$-th ($l > 1$) backbone layer output feature and the previous level discriminator output, so we have: 
\begin{equation}
    \footnotesize
    \textbf{d}^{S}_{l} = Dis_{l}^{S}(f^{S}_{l}(\textbf{I})) \oplus \textbf{d}^{S}_{l-1}),
    \quad
    \textbf{d}^{T}_{l} = Dis_{l}^{T}(f^{T}_{l}(\textbf{I})) \oplus \textbf{d}^{T}_{l-1}),  
\end{equation}
\begin{equation}
    \footnotesize
    \textbf{d}^{\prime S}_{l} = Dis_{l}^{S}(f^{new}_{l}(\textbf{I})) \oplus \textbf{d}^{\prime S}_{l-1}),
    \quad
    \textbf{d}^{\prime T}_{l} = Dis_{l}^{T}(f^{new}_{l}(\textbf{I})) \oplus \textbf{d}^{\prime T}_{l-1}).   
\end{equation}
After obtaining the output from the last-level discriminator, we define $\Lagr_{D_{S}}$ and $\Lagr_{D_{T}}$ to train $Dis_{s}$ and $Dis_{t}$, and $\Lagr_{S}$ and $\Lagr_{T}$ to supervise $f^{new}$.
\begin{equation}
    \small
    \Lagr_{S} = - \mathbb{E}_{x_{p}\sim P_{s}}[log(\mathbf{d}^{S}_{l})] - \mathbb{E}_{x_{f}\sim P_{new}}[log(1-\textbf{d}^{\prime S}_{l})],
\end{equation}
\begin{equation}
    \small
    \Lagr_{T} = - \mathbb{E}_{x_{p}\sim P_{t}}[log(\mathbf{d}^{T}_{l})] - \mathbb{E}_{x_{f}\sim P_{new}}[log(1-\textbf{d}^{\prime T}_{l})],
\end{equation}
\begin{equation}
    \small
    \Lagr_{D_{s}} = - \mathbb{E}_{x_{p}\sim P_{s}}[log(1 - \mathbf{d}^{S}_{l})] - \mathbb{E}_{x_{f}\sim P_{new}}[log(\textbf{d}^{\prime S}_{l})],
\end{equation}
\begin{equation}
    \small
    \Lagr_{D_{t}} = - \mathbb{E}_{x_{p}\sim P_{t}}[log(1 - \mathbf{d}^{T}_{l})] - \mathbb{E}_{x_{f}\sim P_{new}}[log(\textbf{d}^{\prime T}_{l})].
\end{equation}

\Paragraph{Discussion} The idea of adopting adversarial training on the feature map for knowledge transfer is similar to \cite{chung2020feature}. However, the method in \cite{chung2020feature} is for the online task and transferring knowledge from two models specialized in the same domain. Conversely, our case is to learn from heterogeneous models which specialize in different domains. 
Additionally, using two regularization terms with symmetry based on the two pre-trained models, is similar to work in \cite{zhang2020class} on the knowledge distillation topic that is different to FAS. 
However, the same is the effect of alleviating the imbalance between classification loss and regularization terms, as reported in \cite{kim2021split,zhang2020class}.

%% file: sections/03_04_training_inference.tex
\SubSection{Training and Inference}
\label{sec_training_and_inference}
Our training procedure contains two stages, as depicted in Fig.~\ref{fig:overall_archi}. Firstly, we fine-tune given any source pre-trained FAS model with the proposed SRE, on the target dataset. 
We optimize the model by minimizing the $\ell_1$ distance (denoted as $\Lagr_{Mask}$) between the predicted binary mask $\mathbf{M}$ and $\textbf{I}_{pre}$, and the original loss $\Lagr_{Orig}$ that is used in the training procedure of original FAS models. After the fine-tuning process, we obtain well-trained \textit{SRE} and a feature extractor ($f^{T}$) that is able to work reasonably well on target domain data. Secondly, we integrate the well-trained \textit{SRE} with the updated model ($f^{new}$) and the source pre-trained model ($f^{S}$), such that we can obtain estimated spoof cues from perspectives of two models. We use $\Lagr_{Spoof}$ to prevent the divergence between spoof regions estimated from $f^{new}$ and $f^{S}$. Lastly, we use $\Lagr_{S}$ and $\Lagr_{T}$ as introduced in Sec.~\ref{sec_multi_dis} for transferring knowledge from the $f^{S}$ and $f^{T}$ to $f^{new}$, respectively. Therefore, the overall objective function in the training is denoted as $\Lagr_{total}$:  
\begin{equation}
\Lagr_{total} = \lambda_{1} \Lagr_{Orig} + \lambda_{2} \Lagr_{Spoof} + \lambda_{3} \Lagr_{S} + \lambda_{4} \Lagr_{T},
\end{equation}
where $\lambda_{1} {\text -} \lambda_{4}$ are the weights to balance the multiple terms.
In inference, we only keep new feature extract $f^{new}$ and \textit{SRE}, as depcited in Fig.~\ref{fig:overall_archi} \textcolor{red}{(c)}.


%% file: sections/04_benchmark.tex
\input{sections/figure_table_latex/figure_4_dataset}
\Section{FASMD Dataset}
\label{sec_benchmark}
We construct a new benchmark for MD-FAS, termed FASMD, based on SiW~\cite{liu2018learning}, SiW-Mv2~\cite{guo2022multi} \footnote{The original SiW-M~\cite{liu2019deep} is unavailable due to the privacy issue. We release SiW-Mv2 on \href{http://cvlab.cse.msu.edu/siw-m-spoof-in-the-wild-with-multiple-attacks-database.html}{CVLab website} with \href{https://github.com/CHELSEA234/Multi-domain-learning-FAS}{source code}, and details of SiW-Mv$2$ are in the supplementary Sec.~\textcolor{red}{A}} and Oulu-NPU~\cite{OULU_NPU_2017}.
MD-FAS consists of five sub-datasets: dataset A is the source domain dataset, and B, C, D and E are four target domain datasets, which introduce unseen spoof type, new ethnicity distribution, age distribution and novel illumination, respectively. 
The statistics of the FASMD benchmark are reported in Tab.~\ref{tab_FASMD_benchmark}. 

\begin{wraptable}{r}{0.45\textwidth} 
    \scriptsize
    \begin{tabular}{ccc}
    \hline
    \multicolumn{3}{c}{Video Num/Subject Num} \\ \hline
    \multicolumn{1}{c}{Dataset ID} & \multicolumn{1}{c}{Train} & Test \\ \hline    \multicolumn{1}{c}{A (Source)} & \multicolumn{1}{c}{$4,983$/$603$} & $2,149$/$180$ \\ \hline
    \multicolumn{1}{c}{B (New spoof type)} & \multicolumn{1}{c}{$1,392$/$301$} & $383$/$71$ \\ \hline
    \multicolumn{1}{c}{C (New eth. distribution)} & \multicolumn{1}{c}{$1,024$/$360$} & $360$/$27$ \\ \hline
    \multicolumn{1}{c}{D (New age distribution)} & \multicolumn{1}{c}{$892$/$157$} & $411$/$43$ \\ \hline
    \multicolumn{1}{c}{E (New illu. distribution)} & \multicolumn{1}{c}{$1,696$/$260$} & $476$/$40$ \\ \hline
    \end{tabular}
    \caption{\footnotesize The FASMD benchmark. [Keys: eth.=ethnicity, illu.=illumination.]}
    \label{tab_FASMD_benchmark}
\end{wraptable} 

\Paragraph{New spoof type} As illustrated in Fig.~\textcolor{red}{4}, target domain dataset B has novel spoof types that are excluded from the source domain dataset (A). The motivation for this design is, compared with the \textit{print} and \textit{replay} that are prevalent nowadays, other new spoof types are more likely to emerge and cause threats. As a result, given the fact that, five macro spoof types are introduced in SIW-Mv2 (\textit{print}, \textit{replay}, \textit{3D mask}, \textit{makeup} and \textit{partial manipulation attack}), we select one micro spoof type from other three macro spoof types besides \textit{print} and \textit{replay} to constitute the dataset B, which are \textit{Mannequin mask}, \textit{Cosmetic makeup} and \textit{Funny eyes}.

\Paragraph{New ethnicity distribution} In reality, pre-trained FAS models can be deployed to organizations with certain ethnicity distribution (\textit{e.g.}, African American sports club). Therefore, we manually annotate the ethnicity information of each  subject in three datasets, then devise the ethnicity protocol where dataset A has only $1.1$\% African American samples, but this proportion increases to $52.3$\% in dataset C, as depicted in Fig.~\textcolor{red}{5}.

\Paragraph{New age distribution} Likewise, a FAS model that is trained on source domain data full of college students needs to be deployed to the group with a different age distribution, such as a senior care or kindergartens. 
We estimate the age information by the off-the-shelf tool~\cite{rothe2015dex}, and construct dataset D to have a large portion of subjects over $50$ years old, as seen in Fig.~\textcolor{red}{5}.

\Paragraph{New illumination} Oulu-NPU dataset has three different illumination sessions, and we use methods proposed in \cite{zhou2019deep} to estimate the lighting condition for each sample in SIW and SIW-Mv2 datasets. Then we apply $K$-means \cite{macqueen1967some} to cluster them into $K$ groups. For the best clustering performance, we use "eblow method"~\cite{thorndike1953belongs} to decide the value of $K$. We annotate different illumination sessions as \textit{Dark}, three \textit{Front Light}, \textit{Side Light}, and two \textit{Bright Light} (Fig.~\textcolor{red}{4}), then dataset E introduces the new illumination distribution. 

%% file: sections/figure_table_latex/figure_4_dataset.tex
\begin{figure*}[t]
 \centering
 \includegraphics[scale=0.3]{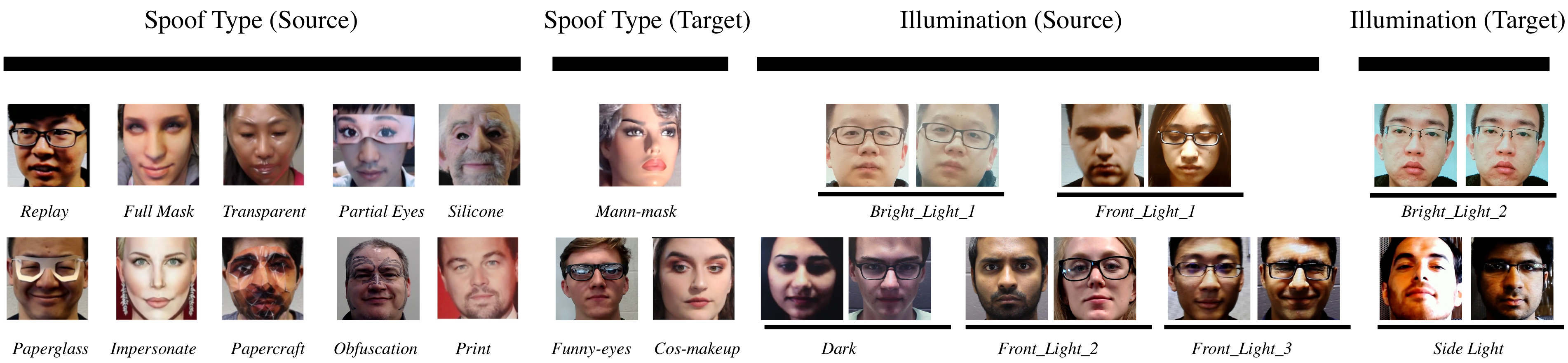}
 \label{fig_dataset_demo}
 \vspace{-3mm}
 \caption{\footnotesize Representative examples in source and target domain for spoof and illumination protocols.}
\end{figure*}
\begin{figure*}[t]
 \centering
 \includegraphics[scale=0.4]{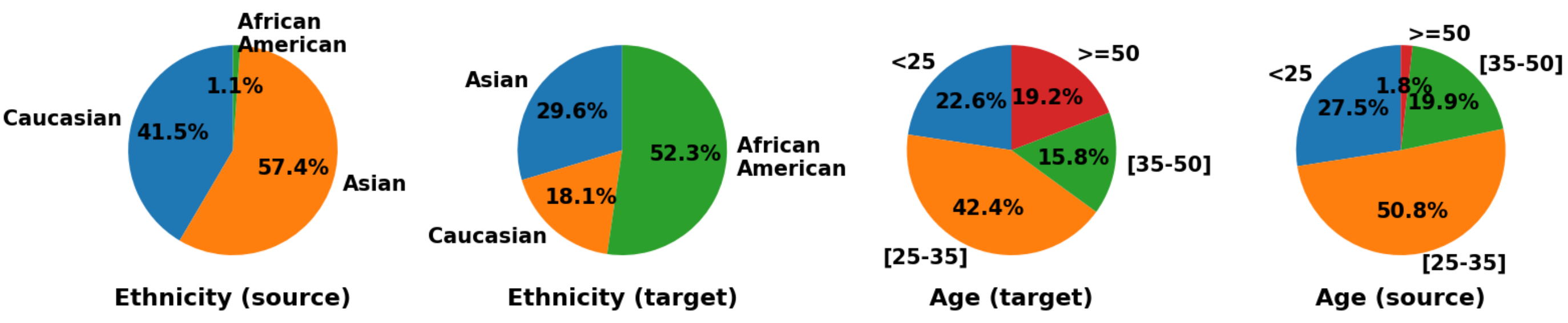}
 \label{fig_distribution}
 \vspace{-3mm}
 \caption{\footnotesize The  distribution of ethnicity and age in source and target domain subsets.  \vspace{-3mm}}
\end{figure*}

%% file: sections/05_00_exp_setup.tex
\section{Experimental Evaluations}
\label{sec_experiment}

\subsection{Experiment Setup}
\label{exp_setup}
We evaluate our proposed method on the FASMD dataset. In Sec.~\ref{sec_adaptability}, we report \textit{FAS-wrapper} performance with different FAS models, and we choose PhySTD~\cite{liu2020physics} as the FAS model for analysis in Sec.~\ref{sec_exp_main}, 
 because PhySTD has demonstrated competitive empirical FAS results.
Firstly, we compare to anti-forgetting methods (\textit{e.g.}, LwF \cite{li2017learning}, MAS \cite{aljundi2018memory} and LwM \cite{dhar2019learning}). Specifically, based on the architecture of PhySTD, we concatenate feature maps generated by last convolution layers in different branches, then employ Global Average Pooling and fully connected (FC) layers to convert concatenated features into a 2-dimensional vector. 
We fix the source pre-trained model weights and only train added FC layers in the original FAS task, as a binary classifier. 
In this way, we can apply methods in~\cite{li2017learning,aljundi2018memory,dhar2019learning} to this binary classifier. 
For multi-domain learning methods (\textit{e.g.}, Serial and Parallel Res-Adapter~\cite{rebuffi2017learning,rebuffi2018efficient}), we choose the $1\times1$ kernel size convolution filter as the adapter and incorporate into the PhySTD as described in original works~(see details in the supplementary material). We use standard FAS metrics to measure the performance, which are Attack Presentation Classification Error Rate (APCER), Bona Fide Presentation Classification Error Rate (BPCER), and Average Classification Error Rate ACER \cite{ACER_source}, Receiver Operating Characteristic (ROC) curve.

\Paragraph{Implementation Details} We use Tensorflow \cite{abadi2016tensorflow} in implementation, and we run experiments on a single NVIDIA TITAN X GPU. In the source pre-train stage, we use a learning rate $3$e-$4$ with a decay rate 0.99 for every epoch and the total epoch number is $180$. We set the mini-batch size as 8, where each mini-batch contains $4$ live images and $4$ spoof images ({\it e.g.}, $2$ SIW-Mv2 images, $1$ image in SIW and OULU-NPU, respectively). Secondly, we keep the same hyper-parameter setting as the pre-train stage, fine-tune the source domain pre-trained model with \textit{SRE} at a learning rate $1$e-$5$. The overall FAS-wrapper is trained with a learning rate $1$e-$6$.

%% file: sections/05_01_experiments.tex
\subsection{Main Results}
\label{sec_exp_main}
\input{sections/figure_table_latex/table_3_main_performance}

Tab. \textcolor{red}{4} reports the detailed performance from different models on all four protocols. 
Overall, our method surpasses the previous best method on source and target domain evaluation in \textit{all categories}.
More importantly, regarding performance on source domain data, it is impressive that our method surpasses the best previous method in all protocols (\textit{e.g.}, $2.5\%$, $1.3\%$, $0.4\%$ and $1.3\%$ for new spoof, enthnicity, age and illumaiton domain respectively, and $0.9\%$ on average). We believe that, the proposed SRE can largely alleviate the \textit{catastrophic forgetting} as mentioned above, thereby yielding the superior source domain performance than prior works. However, the improvement diminishes on the new ethnicity protocol. One possible reason is that the \textit{print} and \textit{replay} attacks account for a large portion of data in new ethnicity distribution, and different methods, performance on these two common presentation attacks are similar.


\input{sections/figure_table_latex/table_4_average_performance}
Additionally, Tab.~\textcolor{red}{5} reports the average performance on four protocols in terms of ACPER, BCPER and ACER. 
Our method still remains the best,
besides BPCER on the target domain performance. It is worth mentioning that we have $9.5\%$ and $5.4\%$ ACER on source and target domain data, respectively, which are better than best results from prior works, namely $5.6\%$ ACER in \cite{dhar2019learning} on target domain data and $10.3\%$ ACER in \cite{rebuffi2018efficient} on source domain data. 
Furthermore, in Sec.~\ref{sec_exp_mul_dis}, we examine the adaptability of our proposed method, by incorporating it with different FAS methods.


\Paragraph{Ablation study using $\Lagr_{Spoof}$} \textit{SRE} plays a key role in \textit{FAS-wrapper} for learning the new spoof type, as ablating the $\Lagr_{Spoof}$ largely decreases the source domain performance, namely from $82.1\%$ to $76.3\%$ on TPR@FPR=$0.5\%$ (Tab. \textcolor{red}{4}) and $0.9\%$ on ACER in source domain~(Tab. \textcolor{red}{5}). Such a performance degradation supports our statement that, $\Lagr_{Spoof}$ prevents divergence between spoof traces estimated from the source teacher and the upgraded model, which helps to combat the \textit{catastrophic forgetting} issue, and maintain the source domain performance.  

\Paragraph{Ablation study using $\Lagr_{S}$ and $\Lagr_{T}$} Without the adversarial learning loss ($\Lagr_{S}$ + $\Lagr_{T}$), the model performance constantly decreases, according to Tab.~\textcolor{red}{4}, although such impacts are less than removal of $\Lagr_{Spoof}$, which still causes $1.3\%$ and $0.7\%$ average performance drop on source and target domains.
Finally, we have a regularization term $\Lagr_{T}$ which also contributes to performance. That is, removing $\Lagr_{T}$ hinders the FAS performance (\textit{e.g.}, $0.4\%$ ACER on target domain performance), as reported in  Tab.~\textcolor{red}{5}.

%% file: sections/figure_table_latex/table_3_main_performance.tex
\begin{table*}[h]
\scriptsize
\centering
\begin{tabular}{c|c|c c c c|c}
\cellcolor[RGB]{222, 164, 151}Method
&\cellcolor[RGB]{222, 164, 151}\begin{tabular}[c]{c}Training data\end{tabular} 
&\cellcolor[RGB]{222, 164, 151}\begin{tabular}[c]{@{}c@{}}Spoof\end{tabular} &\cellcolor[RGB]{222, 164, 151}\begin{tabular}[c]{@{}c@{}}Ethnicity\end{tabular} 
&\cellcolor[RGB]{222, 164, 151}\begin{tabular}[c]{@{}c@{}}Age\end{tabular}&\cellcolor[RGB]{222, 164, 151}\begin{tabular}[c]{@{}c@{}}Illumination\end{tabular}&\cellcolor[RGB]{222, 164, 151}\begin{tabular}[c]{@{}c@{}}Average\end{tabular} \\ \hline

\cellcolor[RGB]{155, 187, 228}Upper Bound & \cellcolor[RGB]{155, 187, 228}Source + Target & \cellcolor[RGB]{155, 187, 228}$92.1/86.5$ & \cellcolor[RGB]{155, 187, 228}$83.4/98.7$ &\cellcolor[RGB]{155, 187, 228}$82.6/86.1$ & \cellcolor[RGB]{155, 187, 228}$83.9/95.4$ & \cellcolor[RGB]{155, 187, 228}$85.5/91.7$\\

\cellcolor[RGB]{155, 187, 228}Source Teacher&\cellcolor[RGB]{155, 187, 228}Source& \cellcolor[RGB]{155, 187, 228}$97.2/63.1$&\cellcolor[RGB]{155, 187, 228}$97.2/86.6$& 
\cellcolor[RGB]{155, 187, 228}$97.2/83.9$& \cellcolor[RGB]{155, 187, 228}$97.2/93.7$& \cellcolor[RGB]{155, 187, 228}$97.2/81.8$\\

\cellcolor[RGB]{155, 187, 228}Target Teacher& \cellcolor[RGB]{155, 187, 228}Target& \cellcolor[RGB]{155, 187, 228}$86.4/88.7$ & \cellcolor[RGB]{155, 187, 228}$74.1/98.7$ & \cellcolor[RGB]{155, 187, 228}$63.0/85.7$ & \cellcolor[RGB]{155, 187, 228}$75.7/95.4$ & \cellcolor[RGB]{155, 187, 228}$74.8/92.0$\\ \hline

LwF \cite{li2017learning} & Target
& $87.6/86.7$ 
& $77.7/98.5$ 
& $71.0/85.2$ 
& $79.5/94.3$ 
& $78.9/91.2$\\

\cellcolor[RGB]{211,211,211}LwM \cite{dhar2019learning} &\cellcolor[RGB]{211,211,211} Target
&\cellcolor[RGB]{211,211,211}$88.8/88.0$ 
&\cellcolor[RGB]{211,211,211}$79.1/98.5$ 
&\cellcolor[RGB]{211,211,211}$73.4/85.1$ 
&\cellcolor[RGB]{211,211,211}$79.9/94.3$
& \cellcolor[RGB]{211,211,211}$80.3/91.5$\\ 

MAS \cite{aljundi2018memory} & Target 
& $88.8/\textcolor{blue}{\textbf{88.1}}$ 
& $79.6/98.2$
& $74.0/\textcolor{blue}{\textbf{85.7}}$
& $80.2/94.4$
& $80.7/91.6$\\

\cellcolor[RGB]{211,211,211} Seri. RA \cite{rebuffi2017learning}
&\cellcolor[RGB]{211,211,211}Target
&\cellcolor[RGB]{211,211,211}$88.3/87.7$
&\cellcolor[RGB]{211,211,211}$\textcolor{blue}{\textbf{80.6}}/97.8$
&\cellcolor[RGB]{211,211,211}$73.6/85.2$
&\cellcolor[RGB]{211,211,211}$79.4/85.1$ 
&\cellcolor[RGB]{211,211,211}$80.5/89.0$\\ 

Para. RA \cite{rebuffi2018efficient} &Target
& $89.0/87.9$ 
& $80.2/98.0$ 
& $73.8/85.2$ 
& $80.1/94.8$ 
& $\textcolor{blue}{\textbf{81.2}}/91.5$\\\hline

\cellcolor[RGB]{211,211,211}\textit{Ours} - $\Lagr_{T}$ &
\cellcolor[RGB]{211,211,211}Target& 
\cellcolor[RGB]{211,211,211}$\textcolor{blue}{\textbf{91.0}}/\textcolor{red}{\textbf{88.4}}$& 
\cellcolor[RGB]{211,211,211}$80.3/98.4$&
\cellcolor[RGB]{211,211,211}$\textcolor{blue}{\textbf{74.0}}/85.6$& 
\cellcolor[RGB]{211,211,211}$80.0/\textcolor{blue}{\textbf{95.3}}$&
\cellcolor[RGB]{211,211,211}$80.8/\textcolor{blue}{\textbf{91.9}}$\\

\begin{tabular}[c]{@{}c@{}}\textit{Ours} - $(\Lagr_{T}+\Lagr_{S})$\end{tabular}&Target&
$90.3/87.3$&
$77.5/\textcolor{red}{\textbf{98.9}}$&
$71.0/85.5$& 
$\textcolor{red}{\textbf{81.5}}/\textcolor{blue}{\textbf{95.3}}$ &
$80.8/91.3$\\

\cellcolor[RGB]{211,211,211}\textit{Ours} - $\Lagr_{Spoof}$&
\cellcolor[RGB]{211,211,211}Target&
\cellcolor[RGB]{211,211,211}$88.4/87.2$&
\cellcolor[RGB]{211,211,211}$75.6/98.3$&
\cellcolor[RGB]{211,211,211}$64.4/85.0$& 
\cellcolor[RGB]{211,211,211}$76.5/95.1$&
\cellcolor[RGB]{211,211,211}$76.3/91.1$\\

\textit{Ours}& Target& 
$\textcolor{red}{\textbf{91.5}}/87.6$ &
$\textcolor{red}{\textbf{81.9}}/\textcolor{blue}{\textbf{98.8}}$ &
$\textcolor{red}{\textbf{74.4}}/\textcolor{red}{\textbf{86.1}}$ & $\textcolor{blue}{\textbf{80.7}}/\textcolor{red}{\textbf{95.4}}$ &
$\textcolor{red}{\textbf{82.1}} / \textcolor{red}{\textbf{92.0}}$\\ \hline

\end{tabular}
\vspace{2mm}
\label{tab_main_performance_sub_dataset}
\caption{\footnotesize The main performance reported in TPR@FPR=$0.5$\%. Scores before and after ``$/$'' are performance on the source and target domains respectively. [Key: \textcolor{red}{\textbf{Best}}, \textcolor{blue}{\textbf{Second Best}}, except for two teacher models and upper bound performance in the first three rows (\textcolor{cell_blue}{\rule{0.4cm}{0.3cm}})].
\vspace{-9mm}}
\end{table*}
%

%% file: sections/figure_table_latex/table_4_average_performance.tex
\begin{wraptable}{r}{0.51\textwidth} 
\vspace{-5mm}
        \scriptsize
        \centering
        \begin{tabular}{c|c|c|c}
        \cellcolor[RGB]{222, 164, 151}Method&\cellcolor[RGB]{222, 164, 151}APCER (\%)&\cellcolor[RGB]{222, 164, 151}BPCER (\%) &\cellcolor[RGB]{222, 164, 151}ACER (\%)\\ \hline

        \cellcolor[RGB]{155, 187, 228} Upper Bound 
        & \cellcolor[RGB]{155, 187, 228}$8.9/4.2$
        & \cellcolor[RGB]{155, 187, 228}$8.0/8.2$ 
        & \cellcolor[RGB]{155, 187, 228}$8.5/6.3$\\

        \cellcolor[RGB]{155, 187, 228}Source Teacher&
        \cellcolor[RGB]{155, 187, 228}$5.0/4.1$ 
        &\cellcolor[RGB]{155, 187, 228}$2.7/18.3$
        &\cellcolor[RGB]{155, 187, 228}$3.8/11.2$\\
        
        \cellcolor[RGB]{155, 187, 228}Target Teacher&
        \cellcolor[RGB]{155, 187, 228}$11.7/5.4$ & 
        \cellcolor[RGB]{155, 187, 228}$10.5/5.1$ & 
        \cellcolor[RGB]{155, 187, 228}$11.1/5.3$\\ \hline
        
        LwF \cite{li2017learning}&
        $11.4/6.6$& 
        $10.1/4.9$ & 
        $10.7/5.7$\\
        
        \cellcolor[RGB]{211,211,211}LwM \cite{dhar2019learning}&
        \cellcolor[RGB]{211,211,211}$9.6/$\textcolor{red}{$\textbf{4.3}$}&
        \cellcolor[RGB]{211,211,211}$11.2/7.0$&
        \cellcolor[RGB]{211,211,211}$10.4/5.6$\\
        
        MAS \cite{aljundi2018memory}
        &$11.7/6.3$
        &$10.4/6.9$
        &$11.5/6.6$\\
        
        \cellcolor[RGB]{211,211,211}Seri. RA \cite{rebuffi2017learning}&
        \cellcolor[RGB]{211,211,211}$10.0/5.0$&
        \cellcolor[RGB]{211,211,211}$11.3/9.0$&
        \cellcolor[RGB]{211,211,211}$10.6/7.0$\\ 
        
        Para. RA \cite{rebuffi2018efficient} 
        &$10.7$/\textcolor{blue}{$\textbf{4.6}$}
        &$9.8/8.6$
        &$10.3/6.6$\\ \hline
        
        \cellcolor[RGB]{211,211,211}\textit{Ours} - $\Lagr_{T}$
        &\cellcolor[RGB]{211,211,211}$9.8/5.2$
        &\cellcolor[RGB]{211,211,211}$9.9/6.4$
        &\cellcolor[RGB]{211,211,211}\textcolor{blue}{$\textbf{9.9}$}/\textcolor{blue}{$\textbf{5.8}$}\\

        \textit{Ours} - ($\Lagr_{S}$ + $\Lagr_{T}$)&
        $10.2/4.8$&
        \textcolor{blue}{$\textbf{9.8}$}/$7.0$&
        $10.0$/\textcolor{red}{$\textbf{5.4}$}\\        
        
        \cellcolor[RGB]{211,211,211}\textit{Ours} - $\Lagr_{Spoof}$&
        \cellcolor[RGB]{211,211,211}$10.3/6.0$&
        \cellcolor[RGB]{211,211,211}$10.6$/\textcolor{red}{$\textbf{5.8}$}&
        \cellcolor[RGB]{211,211,211}$10.4$/5.9\\
        
        \textit{Ours}&
        $9.4/9.5$&
        \textcolor{red}{$\textbf{9.5}$}/\textcolor{blue}{$\textbf{5.9}$}& \textcolor{red}{$\textbf{9.5}$}/\textcolor{red}{$\textbf{5.4}$}\\
        
        \hline                           
        \end{tabular}
    \label{tab_main_performance_average}
    \vspace{-1mm}
    \caption{\footnotesize The average performance of the different methods in four protocols. The scores before and after ``$/$'' are performance on source and target domains. [Key: \textcolor{red}{\textbf{Best}}, \textcolor{blue}{\textbf{Second Best}}, except for two teacher models and upper bound performance in first three rows (\textcolor{cell_blue}{\rule{0.4cm}{0.3cm}})].}
\end{wraptable} 

%% file: sections/05_02_adaptability.tex
\subsection{Adaptability Analysis}
\label{sec_adaptability}
\input{sections/figure_table_latex/table_5_combined_table}
\input{sections/figure_table_latex/figure_spoof_region_plugin}
We apply \textit{FAS-wrapper} on three different FAS methods: Auxi.-CNN~\cite{liu2018learning}, CDCN~\cite{yu2020searching} and PhySTD~\cite{liu2020physics}. CDCN uses a special convolution (\textit{i.e.}, Central Difference Convolution) and Auxi.-CNN is the flagship work that learns FAS via auxiliary supervisions. As shown in Tab.~\ref{tab_combine_table}~\textcolor{red}{(a)}, \textit{FAS-wrapper} can consistently improve the performance of naive fine-tuning. When ablating the $\Lagr_{Spoof}$, PhySTD \cite{liu2020physics} experiences the large performance drop ($5.8\%$) on the source domain, indicating the importance of SRE in the learning the new domain. Likewise, the removal of adversarial learning loss (\textit{e.g.}, $\Lagr_{T} + \Lagr_{S}$) leads to difficulty in preserving the source domain performance,
which can be shown from, on the source domain, CDCN~\cite{yu2020searching} decreases $1.4\%$ and Auxi-CNN~\cite{liu2018learning} decreases $3.0\%$. This means dual teacher models, in the \textit{FAS-wrapper}, trained with adversarial learning benefit the overall FAS performance.
Also, we visualize the spoof region generated from \textit{SRE} with \cite{liu2018learning} and \cite{liu2020physics}
in Fig.~\ref{fig_reflection}. 
We can see the spoof cues are different, which supports our hypothesis that, although FAS models make the same final binary prediction, they internally identify spoofness in different areas. 

%% file: sections/figure_table_latex/table_5_combined_table.tex
\begin{table}[t]
\begin{minipage}[b]{0.5\linewidth}
\footnotesize
\centering
    \resizebox{0.8\textwidth}{!}{
    \begin{tabular}{cccc}
    \hline
    \begin{tabular}[c]{@{}c@{}}TPR@FPR=0.5\%\\ (Source/Target)\end{tabular} & PhySTD~\cite{liu2020physics} & CDCN~\cite{yu2020searching} & Auxi.-CNN~\cite{liu2018learning} \\ \hline
    Naive Fine. & 
    $74.8/92.0$&$72.2/91.2$&$70.9/90.8$ \\ \hline
    
    \textit{Full} (Ours)& 
    $82.1/92.0$&$80.0/91.4$&$78.0/91.8$ \\ \hline
    
    \textit{Full} - $\Lagr_{Spoof}$&
    $76.3/91.1$&$76.1/91.5$&$69.0/91.0$\\ \hline
    
    \textit{Full} - $\Lagr_{T}$&
    $81.3/91.9$&$79.9/91.3$&$78.0/90.6$\\ \hline
    
    \textit{Full} - $(\Lagr_{T}+\Lagr_{S})$&
    $80.1/91.3$&$78.6/92.1$&$75.0/91.3$\\ \hline
    \end{tabular}}
\end{minipage}\hfill
\begin{minipage}[b]{0.5\linewidth}
\footnotesize
\centering
    \resizebox{0.9\textwidth}{!}{
    \begin{tabular}{cc}
    \hline
     & TPR@FPR=0.5\% \\ \hline
    $\Lagr_{Spoof}$ + Multi-disc. (Ours) & $82.1/92.0$ \\ \hline
    $\Lagr_{Spoof}$ + Multi-disc. (same weights) & $79.0/91.8$ \\ \hline
    $\Lagr_{Spoof}$ + Single disc. (concat.)& $78.4/91.6$ \\ \hline
    $\Lagr_{Spoof}$ + \cite{tung2019similarity} & $75.1/91.7$ \\ \hline
    \end{tabular}}
\end{minipage}
\vspace{1mm}
\caption{\footnotesize (a) The \textit{FAS-wrapper} performance with different FAS models; (b) Performance of adopting different architecture design choices. \vspace{-5mm}}
\label{tab_combine_table}
\end{table}

%% file: sections/figure_table_latex/figure_spoof_region_plugin.tex
\begin{figure}[t]
 \centering
 \includegraphics[scale=0.26]{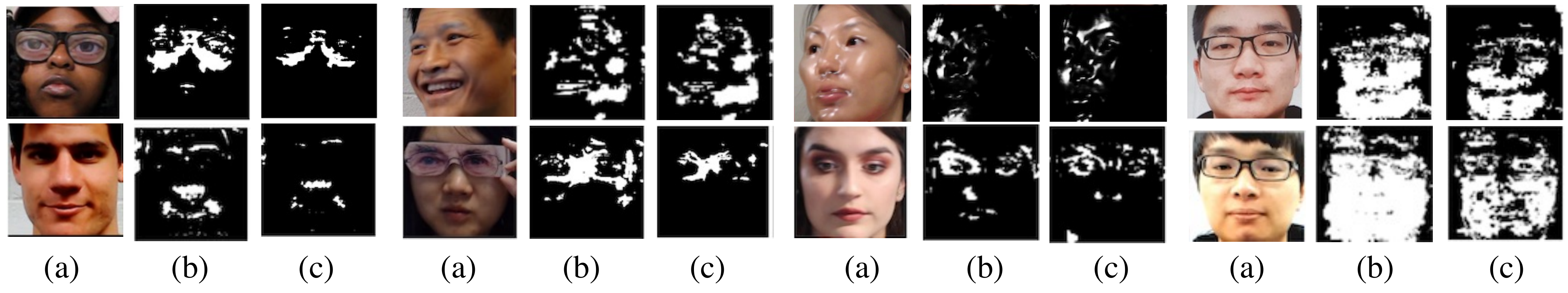}
 \vspace{-7mm}
 \caption{\footnotesize Spoof region estimated from different models. Given input image (a), (b) and (c) are model responses from \cite{liu2020physics} and \cite{liu2018learning}, respectively. Detailed analyses in Sec.~\ref{sec_exp_sp}.
  \vspace{-5mm}}
\label{fig_reflection}
\end{figure}

%% file: sections/05_03_exp_sp.tex
\subsection{Algorithm Analysis}
\label{sec_exp_sp}
\Paragraph{Spoof Region Visualization} We feed output features from different models (\textit{i.e.}, $f^{S}$, $f^{T}$ and $f^{new}$) to a well-trained \textit{SRE} to generate the spoof region, as depicted in Fig.~\ref{fig_pre_fin_model_response}.
In general, the $f^{S}$ produces more accurate activated spoof regions on the source domain images. For example, two source images in new spoof category have detected makeup spoofness on eyebrows and mouth (first row) and more intensive activation on the funny eye region (second row). $f^{T}$ has the better spoof cues estimated on the target domain image. For example, two target images in the new spoof category, where spoofness estimated from $f^{T}$ is stronger and more comprehensive; in the novel ethnicity category, the spoofness covers the larger region. With $\Lagr_{Spoof}$, the updated model ($f^{new}$) identifies the spoof traces in a more accurate way.
\input{sections/figure_table_latex/figure_spoof_region_pretrain_finetune}
\input{sections/figure_table_latex/figure_spoof_region_mask}

\Paragraph{Explanability} We compare \textit{SRE} with the work which generate binary masks indicating the spoofness \cite{on-the-detection-of-digital-face-manipulation}, and works which explain how a model makes a binary classification decision~\cite{zhou2016learning,selvaraju2017grad}. In Fig.~\ref{fig_explain}, we can observe that our generated spoof traces can better capture the manipulation area, regardless of spoof types. For example, in the first \textit{print} attack image, the entire face is captured as spoof in our method but other three methods fail to achieve so. Also, 
our binary mask is more detailed and of higher resolution than that of \cite{on-the-detection-of-digital-face-manipulation}, and more accurate and robust than \cite{zhou2016learning,selvaraju2017grad}. Notably, we do not include works in \cite{liu2020physics,liu2020disentangling,zhao2021multi} which use many outputs to identify spoof cues.

\Paragraph{Architecture Design} We compare to some other architecture design choices, such as all multi-scale discriminators with the same weights, concatenation of different scale features and one single discriminator. Moreover, we use correlation similarity table in \cite{tung2019similarity} instead of multi-scale discriminators for transfering knowledge from $f^{S}$ and $f^{T}$ to $f^{new}$. Tab.~\ref{tab_combine_table}~\textcolor{red}{(b)} demonstrates the superiority of our architectural design.

%% file: sections/figure_table_latex/figure_spoof_region_pretrain_finetune.tex
\begin{figure*}[t]
 \centering
 \includegraphics[scale=0.23]{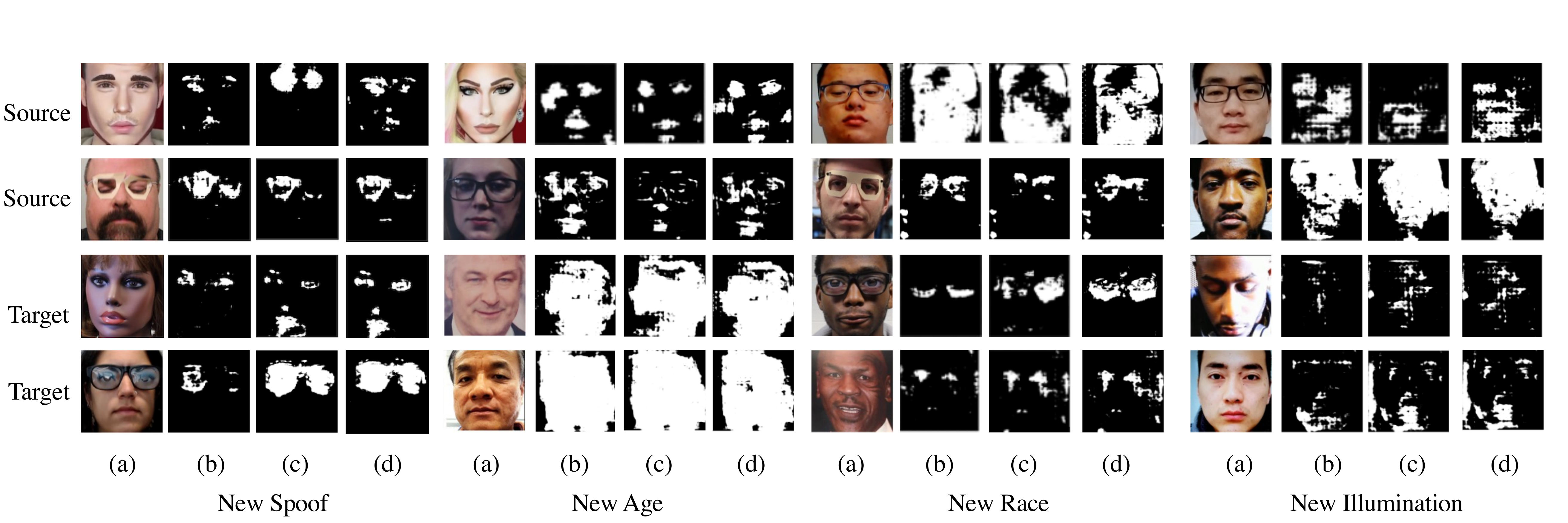}
 \vspace{-0.8cm}
 \caption{\footnotesize Given the input spoof image (a), spoof regions generated by \textit{SRE} with two teacher models (\textit{i.e.}, $f^{S}$ and $f^{T}$) in (b) and (c), and the new upgraded model ($f^{new}$) in (d), for different protocols. Detailed analyses are in Sec.~\ref{sec_exp_sp}.}
 \label{fig_pre_fin_model_response}
\end{figure*}
\vspace{-1mm}

%% file: sections/figure_table_latex/figure_spoof_region_mask.tex
\begin{figure}[t]
 \centering
 \includegraphics[scale=0.4]{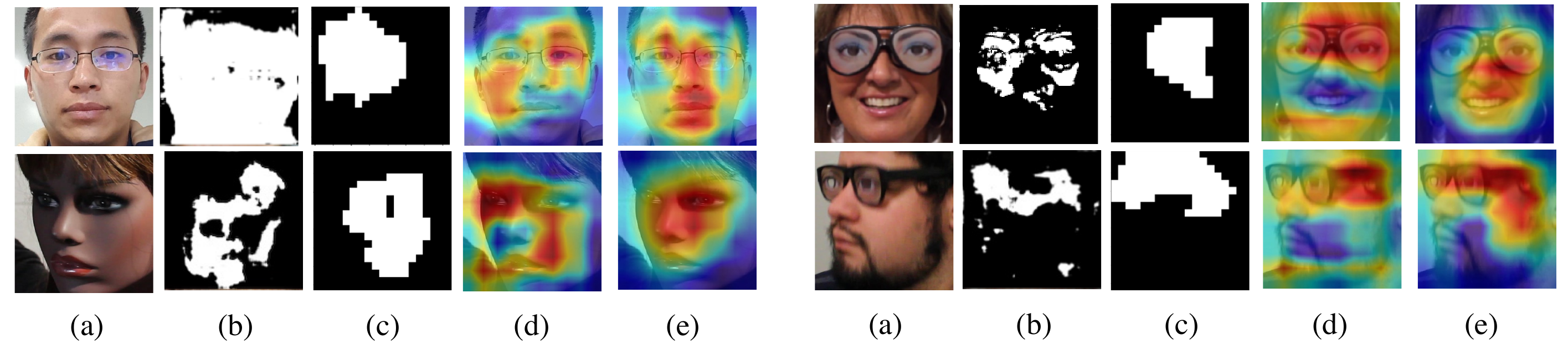}
 \vspace{-3mm}
 \caption{\footnotesize Different spoof estimate methods. Given input image (a), (b) and (c) are the spoof regions estimated from ours and \cite{on-the-detection-of-digital-face-manipulation}. (d) and (e) are the activated map from methods in \cite{selvaraju2017grad} and \cite{zhou2016learning}.\vspace{-5mm}}
 \label{fig_explain}
\end{figure}

%% file: sections/05_04_exp_dual_models.tex
\label{sec_exp_mul_dis}

%% file: sections/05_05_cross_domain_exp.tex
\vspace{1mm}
\subsection{Cross-dataset Study}
\label{sec_cross_dataset}
\begin{figure*}[t!]
 \centering
 \includegraphics[scale=0.3]{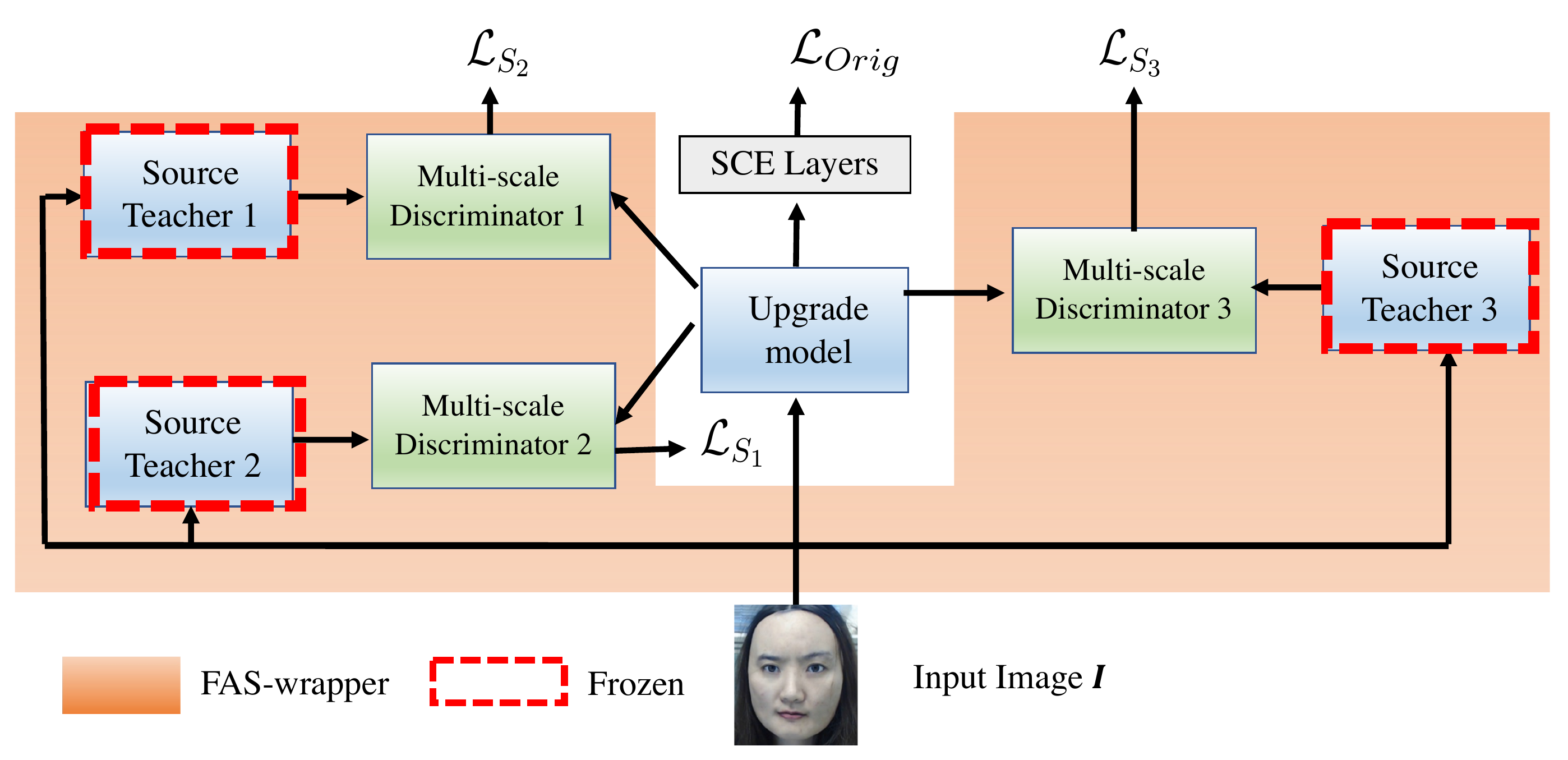}
 \vspace{-4mm}
 \caption{\footnotesize We adapt \textit{FAS-wrapper} for the cross-dataset scenario. \vspace{-1mm}}
 \label{fig_tsne}
\end{figure*}

We evaluate our methods in the cross-dataset scenario and compare to SSDG~\cite{jia2020single} and MADDG~\cite{shao2019multi}. Specifically, we denote OULU-NPU \cite{OULU_NPU_2017} as O, SIW \cite{liu2018learning} as S, SIW-Mv2 \cite{liu2019deep} as M, and HKBU-MARs \cite{liu20163d} as H. 
We use three datasets as source domains for training and one remaining dataset for testing. 
We train three individual source domain teacher models on three source datasets respectively. 
Then, as depicted in Fig.~\ref{fig_tsne}, inside \textit{FAS-wrapper}, three multi-scale discriminators are employed to transfer knowledge from three teacher models to the updated model $f^{new}$ which is then evaluated on the target domain. 
Notably, we remove proposed \textit{SRE} in this cross-dataset scenario, as there is no need to restore the prior model responses. 
\begin{table}
    \tiny
    \scriptsize
    \centering
    \resizebox{1\linewidth}{!}{
    \begin{tabular}{c|cc|cc|cc|cc}
        \hline
        & \multicolumn{2}{c|}{\textbf{O\&M\&H to S}} & \multicolumn{2}{c|}{\textbf{O\&W\&H to M}} & \multicolumn{2}{c|}{\textbf{M\&S\&H to O}} & \multicolumn{2}{c}{\textbf{M\&S\&O to H}} \\ \hline
        
        & \multicolumn{1}{l|}{HTER(\%)} & AUC(\%)& \multicolumn{1}{l|}{HTER(\%)} & AUC(\%)& \multicolumn{1}{l|}{HTER(\%)} & AUC(\%)& \multicolumn{1}{l|}{HTER(\%)} & AUC(\%)\\ \hline
        
        MADDG~\cite{shao2019multi} & \multicolumn{1}{c|}{$16.7$} & $90.5$ & \multicolumn{1}{c|}{$50.3$} & $60.7$ &  \multicolumn{1}{c|}{$17.6$} & $73.0$  &  \multicolumn{1}{c|}{$33.2$} & $73.5$  \\ \hline
        
        SSDG-M~\cite{jia2020single} & \multicolumn{1}{c|}{$\textbf{11.1}$} & $93.4$ & \multicolumn{1}{c|}{$29.6$} & $67.1$ &  \multicolumn{1}{c|}{$\textbf{12.1}$} & $\textbf{89.0}$ & \multicolumn{1}{c|}{$\textbf{25.0}$} & $82.5$  \\ \hline
        
        SSDG-R~\cite{jia2020single} & \multicolumn{1}{c|}{$13.3$} & $93.4$ &  \multicolumn{1}{c|}{$29.3$} & $\textbf{69.5}$  &  \multicolumn{1}{c|}{$13.3$} & $83.4$  &  \multicolumn{1}{c|}{$28.9$} & $81.0$  \\ 
        \hline
        \hline
        
        Ours &  \multicolumn{1}{c|}{$15.4$} & $\textbf{93.6}$  &  \multicolumn{1}{c|}{$\textbf{28.1}$} & $68.4$  &  \multicolumn{1}{c|}{$14.8$}&$85.6$  &  
        \multicolumn{1}{c|}{$27.1$} & $\textbf{83.8}$ \\ \hline
    \end{tabular}
    }
    \label{tab_cross_dataset_performance}
    \vspace{1mm}
    \caption{\footnotesize The cross-dataset comparison.}
\end{table}

The results are reported in Tab.~\textcolor{red}{7}, indicating that our \textit{FAS-wrapper} also exhibits a comparable performance on the cross-dataset scenario as prior works.

%% file: sections/06_conclusion.tex
\vspace{-2mm}
\section{Conclusion}
\vspace{-2mm}
We study the multi-domain learning face anti-spoofing (MD-FAS), which requires the model perform well on both source and novel target domains, after updating the source domain pre-trained FAS model only with target domain data. We first summarize the general form of FAS models, then based on which we develop a new architecture, \textit{FAS-wrapper}. \textit{FAS-wrapper} contains spoof region estimator which identifies the spoof traces that help combat \textit{catastrophic forgetting} while learning new domain knowledge, and the \textit{FAS-wrapper} exhibits a high level of flexibility, as it can be adopted by different FAS models. The performance is evaluated on our newly-constructed FASMD benchmark, which is also the first MD-FAS dataset in the community.

\vspace{1mm}
\Paragraph{Acknowledgment} This research is based upon work supported by the Office of the Director of National Intelligence (ODNI), Intelligence Advanced Research Projects Activity (IARPA), via IARPA R\&D Contract No. $2017$-$17020200004$. The views and conclusions contained herein are those of the authors and should not be interpreted as necessarily representing the official policies or endorsements, either expressed or implied, of the ODNI, IARPA, or the U.S. Government. The U.S. Government is authorized to reproduce and distribute reprints for Governmental purposes notwithstanding any copyright annotation thereon. 





%% file: sections/07_supplementary.tex
\newpage
\section*{Supplementary}
\renewcommand{\thesection}{\Alph{section}}
\setcounter{section}{0}

\input{figures_supp/siw_m_gallery_0}

\noindent In this supplementary material, we provide:

$\diamond$ The detailed introduction of the SiW-Mv$2$ dataset and its designed protocols.

$\diamond$ The baseline model and its performance on the SiW-Mv$2$ dataset.

$\diamond$ Additional explanations of the proposed FAS-\textit{wrapper}.

$\diamond$ Additional implementation details of our experiments in Sec.~\ref{sec_exp_main}.

\Section{SiW-Mv$2$ Dataset}\label{sec_siwm}
\input{figures_supp/siw_m_gallery_1}
In this section, Sec.~\ref{sec_siwm_intro} and \ref{sec_protocol} report the SiW-Mv$2$ dataset and three protocols. In Sec.~\ref{sec_performance}, we first verify the baseline performance on the  Oulu-NPU and SiW datasets, and then report the baseline performance on the SiW-Mv$2$ dataset~\footnote{The source code and download instructions can be found on \href{https://github.com/CHELSEA234/Multi-domain-learning-FAS}{this page}.}.

\SubSection{Introduction}\label{sec_siwm_intro}
Our SiW-Mv$2$ dataset is the updated version of the original SiW-M dataset~\cite{liu2019deep}, which is unavailable due to the privacy issue. For the SiW-Mv$2$ dataset, we curate new samples and add one more spoof category (\textit{e.g.}, \textit{partial mouths}) to improve the overall spoof attack diversity. As a result, SiW-Mv$2$ has $785$ videos from $493$ live subjects, and $915$ spoof videos from $600$ subjects. Among these spoof videos, we have $14$ spoof attack types, spanning from typical $2$D spoof attacks (\textit{e.g.}, \textit{print} and \textit{replay}), various masks, different makeups, and physical material coverings. Some samples are shown in  Fig.~\ref{fig_gallery} and Fig.~\ref{fig_gallery_v2}, and more details are in Tab.~\ref{tab_siwmv2}. 
Moreover, in the SiW-Mv$2$ dataset, spoof attacks can either modify the subject appearance to impersonate other people, such as \textit{impersonate makeup} and \textit{silicone head}, or hide the subject identity (\textit{e.g.}, \textit{funny eyes} and \textit{paper mask}). The details of the dataset collection process are in Sec.~\textcolor{red}{4} of the work~\cite{liu2019deep}. Lastly, the recent usage of SiW-Mv$2$ is also found in the domain of image forensics~\cite{malp,hifi_net}, where methods are developed to distinguish real images from images that are manipulated or generated by Artificial Intelligence.

\input{figures_supp/siw_m_stats}
\SubSection{Protocols and Metrics}\label{sec_protocol}
In the SiW-Mv$2$ dataset, we design three different protocols which evaluate the model ability to detect known and unknown spoof attacks, as well as the generalization ability to spoof attacks at different domains, respectively.
\begin{itemize}
    \item \textbf{Protocol I}: \textit{Known Spoof Attack Detection}. We divide live subjects and subjects of each spoof pattern into train and test splits. We train the model on the training split and report the overall performance on the test split.
    \item \textbf{Protocol II}: \textit{Unknown Spoof Attack Detection}. We follow the leave-one-out paradigm --- keep $13$ spoof attack and $80\%$ live subjects as the train split, and use the remaining one spoof attacks and left $20\%$ live subjects as the test split. We report the test split performance for both individual spoof attacks, as well as the  averaged performance with standard deviation.  
    \item \textbf{Protocol III}: \textit{Cross-domain Spoof Detection}. We partition the SiW-Mv$2$ into $5$ sub-datasets, as described in Sec.~\textcolor{red}{4} in the main paper. We train the model on the source domain dataset, and evaluate the model on test splits of $5$ different domains. Each sub-dataset performance, and averaged performance with standard deviation are reported.
\end{itemize}
To be consistent with the previous work, we use standard FAS metrics to measure the SRENet performance. These metrics are Attack Presentation Classification Error Rate (APCER), Bona Fide Presentation Classification Error Rate (BPCER), Average Classification Error Rate (ACER) \cite{ACER_source}, and Receiver Operating Characteristic (ROC) curve, respectively.

\Section{Baseline and Performance}\label{sec_performance}
We present our baseline architecture,  dubbed SRENet, in Fig.~\ref{fig_sreNet_architecture}. Specifically, the proposed SRENet is based on PhySTD~\cite{liu2020physics}. However, compared to the original PhySTD, we make two modifications, which simplify the model and even achieves the better spoof detection performance. First, the original PhySTD generates three different traces to reconstruct both spoof and live counterparts of the given input image, whereas in SRENet we only leverage these three traces to construct the live counterpart of the given input image. Secondly, we integrate the Spoof Region Estimator (SRE) in Sec.~\textcolor{red}{3.2} into the architecture, and this SRE serves as an attention module to help pinpoint the spoof area by the binary mask. Note that the difference between SRENet and FAS-\textit{wrapper} is fundamental: SRENet is a face spoof detection model, whereas FAS-\textit{wrapper} targets at the multi-domain FAS updating. Furthermore, from the architectural perspective, the proposed SRE plays key roles in both SRENet and FAS-\textit{wrapper}.

Empirically, we first verify the effectiveness of the SRENet on the Oulu-NPU and SiW datasets. The performance is reported in Tab.~\ref{tab_oulu_siw}, which shows that our model performance is comparable with that of state-of-the-art methods, such as PhySTD~\cite{liu2020physics} and PatchNet~\cite{wang2022patchnet}. After that, we report the SRENet performance on the three designed protocols of the SiW-Mv$2$ dataset, and results are in Tab.~\ref{tab_siwm_overall}. 
\input{figures_supp/siw_m_baseline}
\clearpage
\input{figures_supp/siw_m_protocol_1}
\input{figures_supp/siw_m_protocol_2}
\Section{More Method Details}\label{sec_more_details}
\begin{figure*}[t]
    \centering
    \vspace{-1mm}
    \includegraphics[scale=0.3]{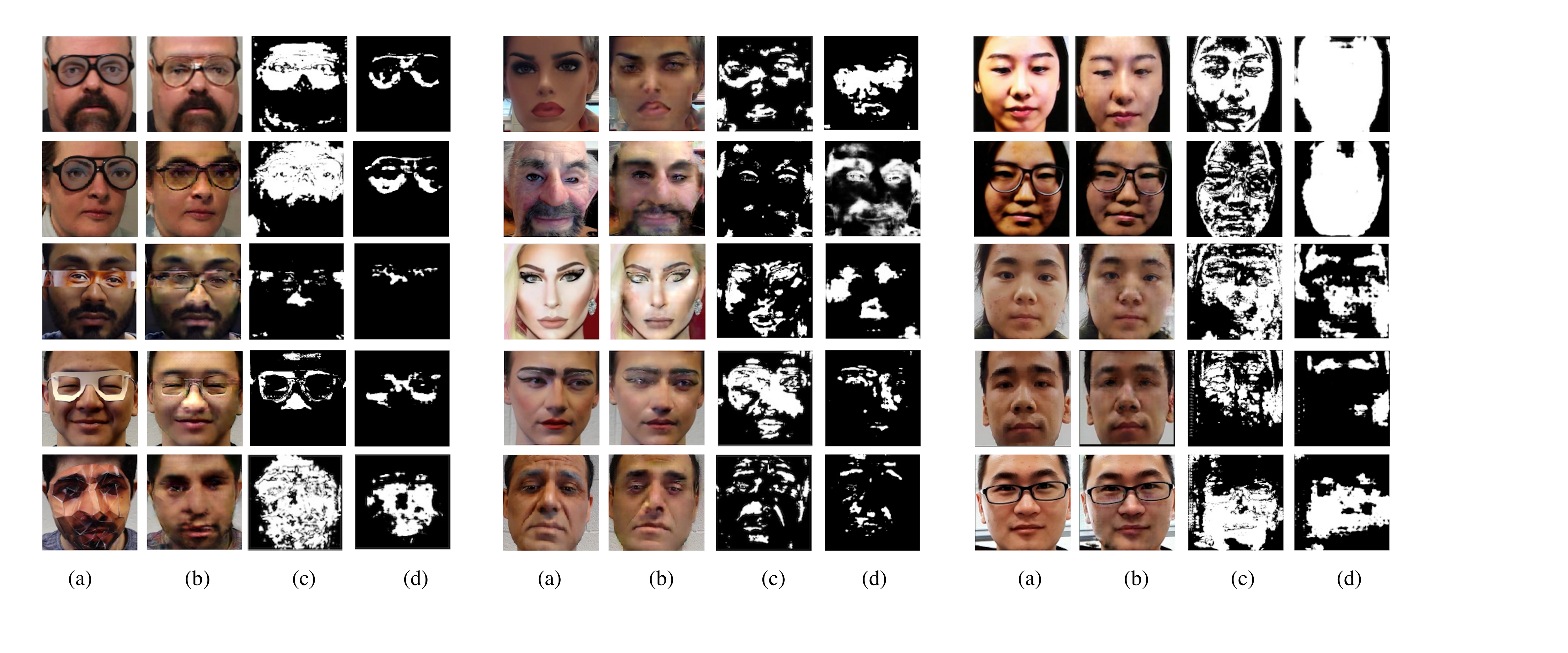}
    \vspace{-0.8cm}
    \caption{\footnotesize The visualization of (a) input spoof image, (b) live counterpart reconstruction, (c) generated preliminary mask, and (d) estimated spoof region from our method. Three columns (from left to right) represent three different spoof genera, covered material, makeup stroke and visual artifacts from \textit{replay} and \textit{print} attack.
    }
    \label{fig_preliminary_mask}
\end{figure*}

\SubSection{Face Reconstruction Analysis}
The motivation of using face reconstruction in our method is that, although it cannot reconstruct the perfect live faces given its spoof counterpart, but it can largely shed the light on spatial pixel location where spoofness occurs. As introduced in Sec.~\textcolor{red}{3.2}, we use such a reconstruction method to generate the preliminary mask $\textbf{I}_{pre}$ as the pseudo label that supervises the proposed \textit{SRE}. We offer the detailed visualization in Fig.~\ref{fig_preliminary_mask}.

In general, we have categorized spoof types into three main genera: (a) covered materials; (b) makeup stroke; (c) visual artifacts (\textit{i.e.}, color distortion and moire effect) in the \textit{replay} and \textit{print} attacks. In particular, for covered materials, reconstruction methods largely erase these spoof materials, such as \textit{funny glasses}, and \textit{paper mask}. As shown from Fig.~\ref{fig_preliminary_mask}, $\textbf{I}_{pre}$ can roughly locate the pixel-wise spatial location that has been covered by the spoof material, and the estimated spoof region gives the more accurate prediction on pixels that are covered by these spoof materials. For makeup stroke, the reconstruction method changes the color and texture of the facial makeup area, making them similar to the natural skin. $\textbf{I}_{pre}$ offers the scattered, discrete binary mask and estimated spoof region provides the smoother region indicating the spoofness. For \textit{replay} and \textit{print} attacks, the reconstruction method modifies facial structure (\textit{i.e.,} nose and eyes) of the human face, or largely change the image's appearance, by providing the image with a sense of depth. Similar as makeup stroke genera, $\textbf{I}_{pre}$ gives very discontinuous predictions on spoofness whereas the estimated spoof region is smoother and semantic.  

\SubSection{Model Response}
When a target domain image $\textbf{I}_{target}$ is fed to the pre-trained model, the pre-trained model will be activated, as if the pre-trained model takes as input source domain images $\textbf{I}_{source}$ which has resemblance with $\textbf{I}_{target}$. In other words, the pre-trained model recognizes it as source domain images $\textbf{I}_{source}$ which has common characteristic and pattern with $\textbf{I}_{target}$. Therefore, source data can manifest themselves on the response of the model, or in other words, keeping model response allows us to have memory or characteristics of the source data.

\begin{figure}[t]
 \centering
 \includegraphics[scale=0.27]{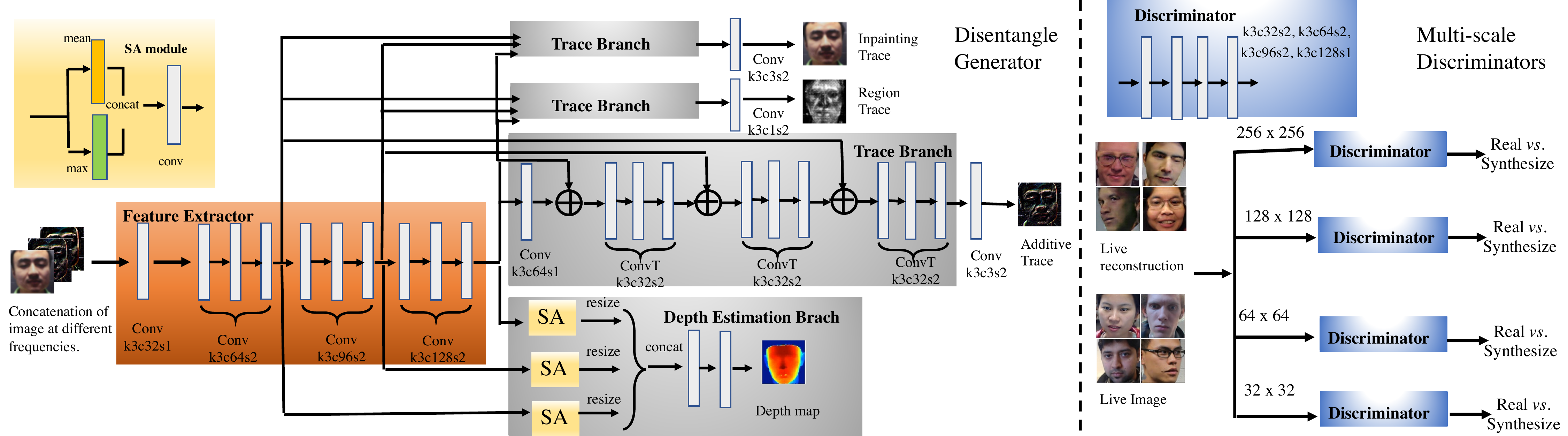}
 \vspace{-3mm}
 \caption{\footnotesize The detailed architecture of PhySTD. The overall architecture contains Disentangle Generator and Multi-scale Discriminators. Notably, in the architecture of PhySTD, each convolutional layer is followed by Batch Normalization layer, RELU activation function and Dropout. This level of details is not included here.}
 \label{fig_pg_architecture}
\end{figure}

\Section{Experimental Implementation}~\label{sec_implementation}
\SubSection{PhySTD method details}
In the experimental section, we apply FAS-\textit{wrapper} on PhySTD for the analysis in Sec.~\textcolor{red}{5.2}. We depict the details of PhySTD in the Fig.~\ref{fig_pg_architecture}, and more can be found in the original work~\cite{liu2020physics}.

\SubSection{The implementation details of prior methods}
We are the first work that studies MD-FAS, in which no source data being available during the model updating process. To the best of our knowledge, there does not exist FAS works in such a source-free scenario. Therefore, in order to have a fair comparison, we need to implement methods from other topics (\textit{e.g.}, \textit{anti-forgetting learning} and \textit{multi-domain learning}) on FAS dataset. In this section, we explain our implementation details on prior methods. 

\begin{figure}[t]
 \centering
 \includegraphics[scale=0.3]{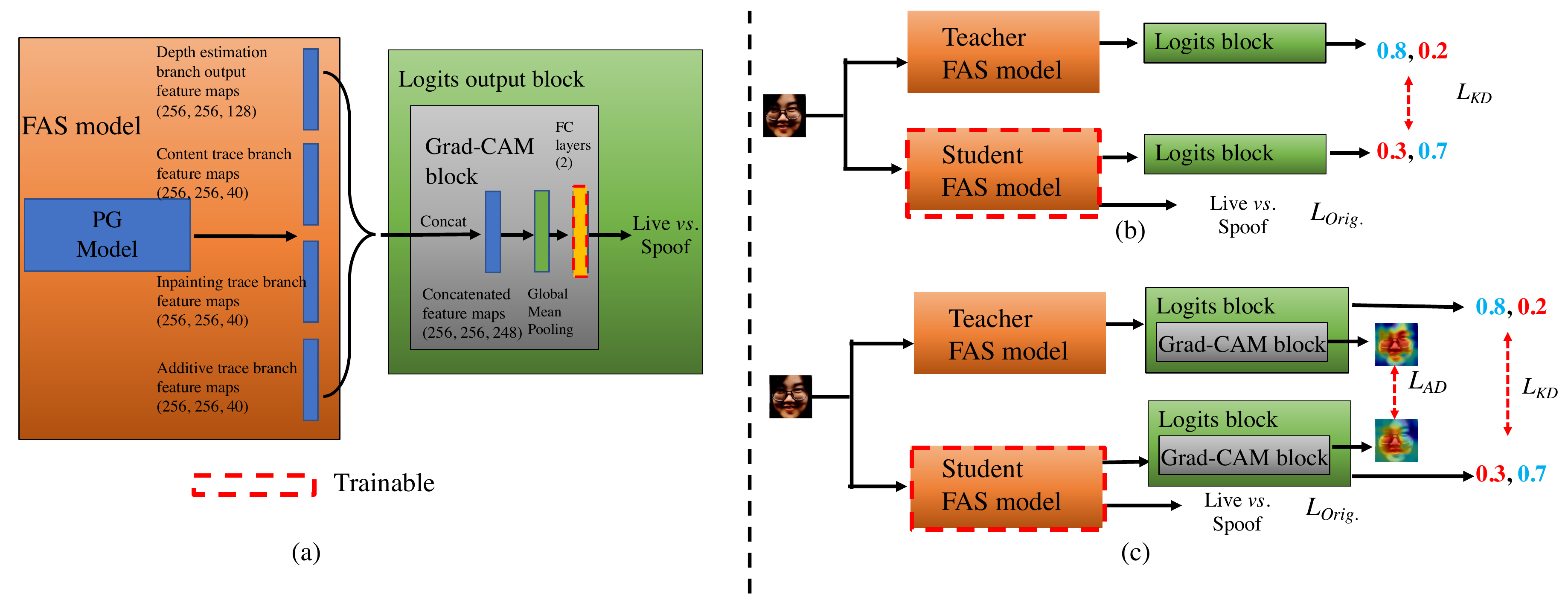}
 \vspace{-3mm}
 \caption{\footnotesize In (a), we modify the pre-trained FAS model into a binary classifier. In (b) and (c), we modified architectures for LwF and LwM methods.}
 \label{fig_prior_anti_forgetting}
\end{figure}

\Paragraph{The implementation details of prior anti-forgetting methods}
We compare our methods to prior works that have anti-forgetting mechanism: LwF \cite{li2017learning}, LwM \cite{dhar2019learning} and MAS \cite{aljundi2018memory}. Firstly, we pre-train the FAS model that is based on PhySTD on the source domain dataset. After the pre-training, we concatenate output feature maps generated from the last convolution layer in different branches as a new concatenated feature maps. 
Then we feed such feature maps through Global Average Pooling Layer and a fully-connected (FC) layer, such that we can obtain a binary classifier. The details are depicted in Fig.~\ref{fig_prior_anti_forgetting}\textcolor{red}{(a)}. We fix the pre-trained FAS model weights and train the last FC layer. As a result, we can use concatenated feature maps for a binary classification result indicating spoofness. We denote newly-added layers as the Logits block, part of which generates the class activate map is denoted as Grad-CAM block. We use these two blocks with the original FAS model for implementing LwF and LwM, as illustrated in Fig.~\ref{fig_prior_anti_forgetting}\textcolor{red}{(b)(c)}. In terms of MAS \cite{aljundi2018memory}, we apply the publicly available source code \footnote{https://github.com/rahafaljundi/MAS-Memory-Aware-Synapses} on the binary classifier we construct, without significantly changing the architecture.
\begin{figure}[t]
 \centering
 \includegraphics[scale=0.35]{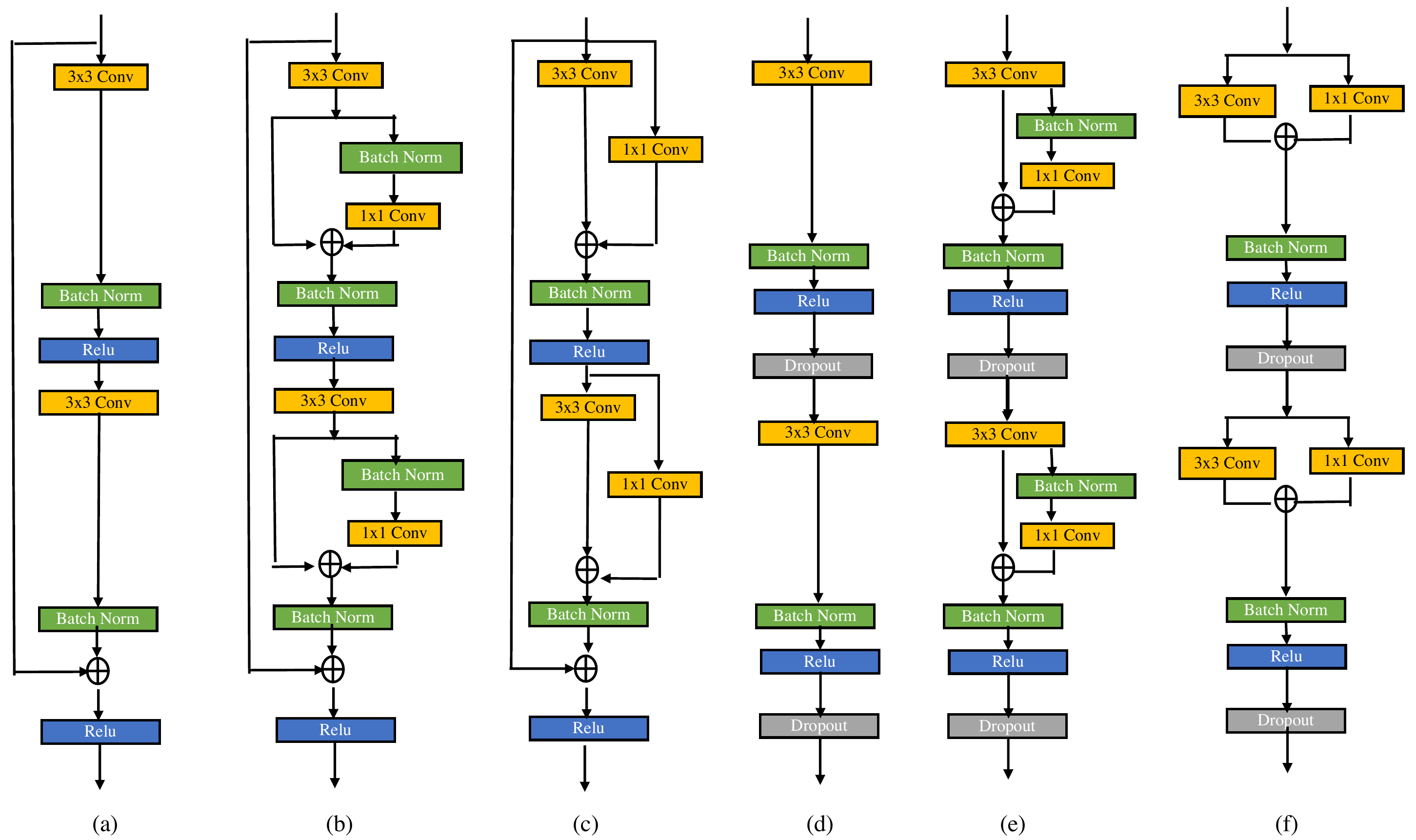}
 \vspace{-3mm}
 \caption{\footnotesize Based on the RseNet building block (a), \cite{rebuffi2017learning,rebuffi2018efficient} have proposed RseNet modified building blocks in (b) and (c) for learning multiple domain knowledge. Likewise, given two consecutive building blocks in PhySTD, we construct modified building blocks, based on \cite{rebuffi2017learning,rebuffi2018efficient}, in (e) and (f). \vspace{-5mm}}
 \label{fig_prior_MDL}
\end{figure}
\Paragraph{The implementation details of multi-domain learning methods}
Seri. Res-Adapter~\cite{rebuffi2017learning}, and Para.~Res-Adapter~\cite{rebuffi2018efficient} are proposed for learning knowledge in multiple visual domains. Specifically, they use domain-specific adapter to enhance model ability in learning a universal image representation for multiple domains. They design such an idea on ResNet~\cite{he2016deep}, which can be seen in Fig.~\ref{fig_prior_MDL}. Based on the same idea, we modify the building block in PhySTD for learning the new domain knowledge. Notably, we have examine different adapter architectures, such as convolution filter with kernel size $1 \times 1$, $3 \times 3$, $5 \times 5$ and $7 \times 7$ , and find that $1 \times 1$ convolution offers the best FAS performance. We also consider the publicly available source code \footnote{https://github.com/srebuffi/residual$\_$adapters} as the reference for the implementation.

%% file: figures_supp/siw_m_gallery_0.tex
\vspace{-4mm}
\begin{figure}
    \centering
    \includegraphics[width=1\linewidth]{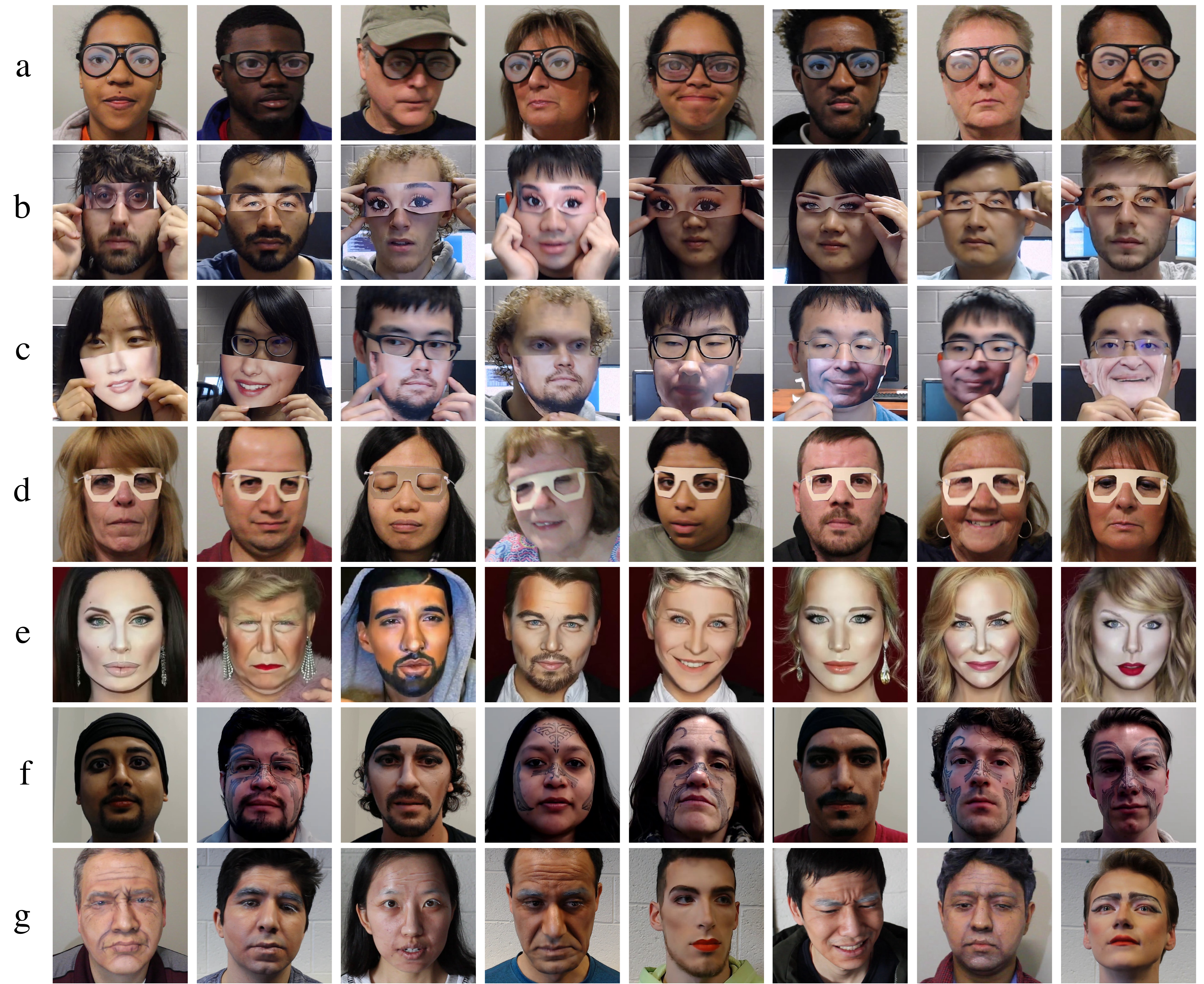}
    \caption{\footnotesize{SiW-Mv$2$ dataset samples in Covering and Makeup spoof categories. These spoof attacks are: (a) Funny Eyes, (b) Partial Eyes, (c) Partial Mouths, (d) Paperglass, (e) Impersonate Makeup, (d) Obfuscation Makeup, and (f) Cosmetic Makeup. More details are in Tab.~\ref{tab_siwmv2}.}}
    \label{fig_gallery}
\end{figure}
\vspace{-1mm}

%% file: figures_supp/siw_m_gallery_1.tex
\begin{figure}[t]
    \centering
    \includegraphics[width=1\linewidth]{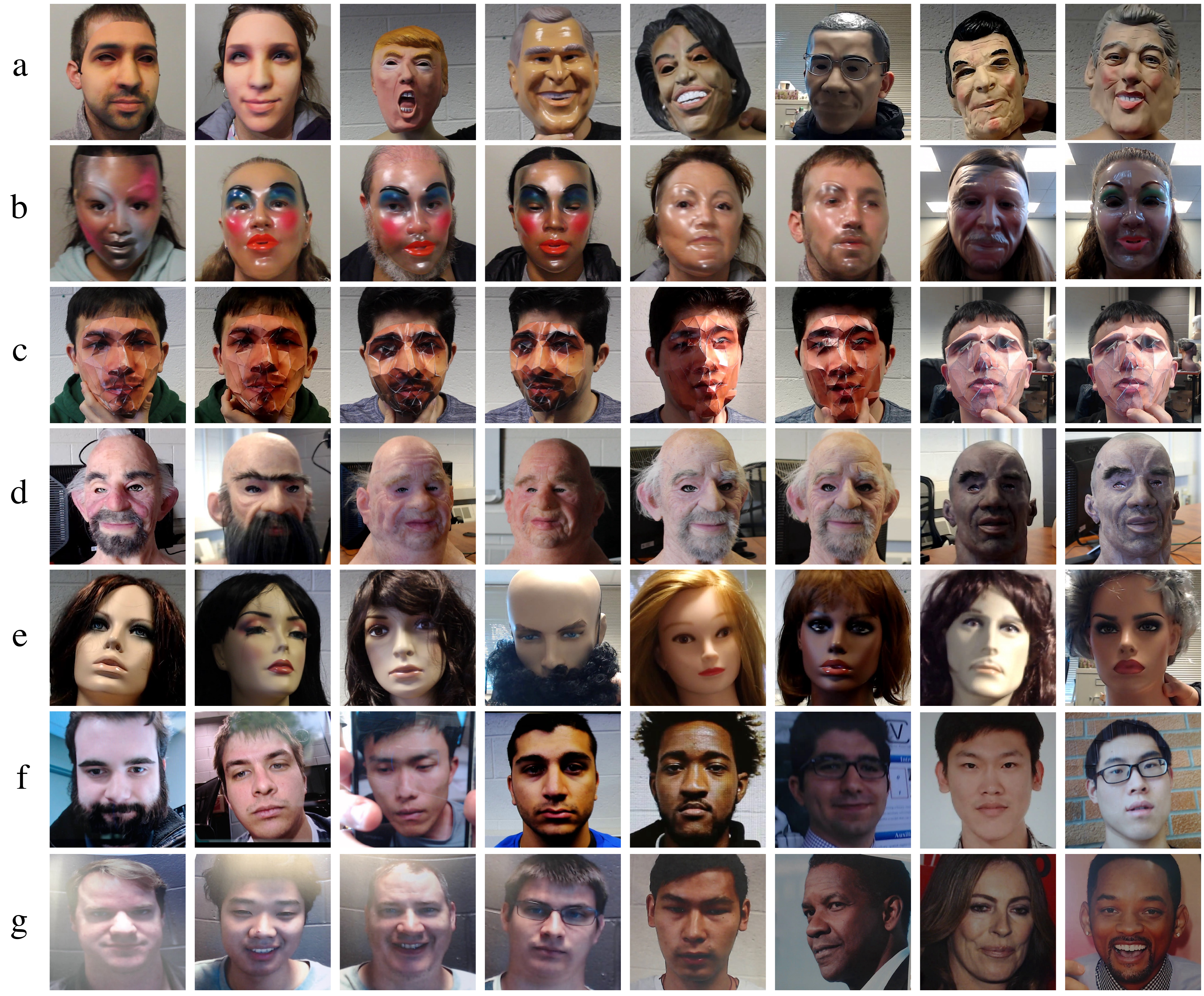}
    \caption{\footnotesize{SiW-Mv$2$ dataset samples in $3$D and $2$D Attack spoof categories. These spoof attacks are: (a) Full Mask, (b) Transparent Mask, (c) Paper Mask, (d) Silicone Head, (e) Mannequin, (d) Print, and (f) Replay. More details are in Tab.~\ref{tab_siwmv2}.}}
    \label{fig_gallery_v2}
\end{figure}

%% file: figures_supp/siw_m_stats.tex
\begin{table}[t]
    \centering
    \small
    \begin{tabular}{c|c|c|c|c}\hline
    Spoof Category&Spoof Attack&Video~$\#$ & Subject~$\#$ & Purpose\\ \hline
    
    \multirow{4}{*}{Covering} & Funny Eyes& $179$ & $172$ & Cover.\\ \cline{2-5} 
        & Partial Eyes & $57$ & $27$ & Hide\\ \cline{2-5} 
        & Partial Mouths & $29$ & $26$ & Hide\\ \cline{2-5} 
        & Paperglasses & $76$ & $71$ & Hide\\ 
        \hline
    
    \multirow{3}{*}{Makeup} & Impersonate & $61$ & $61$ & Imper.\\ \cline{2-5} 
        & Obfuscation & $22$ & $15$ & Imper.\\ \cline{2-5} 
        & Cosmetic & $52$ & $35$ & Imper.\\ 
        \hline
    
    \multirow{5}{*}{$3$D Attack} & Full Mask & $72$ & $12$ & Imper.\\ \cline{2-5}
        & Transparent Mask & $60$ & $60$ & Hide\\ \cline{2-5} 
        & Paper Mask & $17$ & $6$ & Hide\\ \cline{2-5} 
        & Silicone Head & $17$ & $4$ & Hide\\ \cline{2-5} 
        & Mannequin & $40$ & $29$ & Hide\\ 
        \hline
    
    \multirow{2}{*}{$2$D Attack} & Replay& $98$ & $21$ &Imper.\\ \cline{2-5} 
        & Print& $135$ & $61$ &Imper.\\
        \hline
    
    \end{tabular}
    \vspace{1mm}
    \caption{\footnotesize{SiW-Mv$2$ dataset details. Each spoof attack represents a different purpose of spoofing, such as impersonation or hiding the original identity. [\textbf{Keys}: Hide: hiding identity, Imper.: impersonation]\vspace{-3mm}}}
    \label{tab_siwmv2}
\end{table}

%% file: figures_supp/siw_m_baseline.tex
\begin{figure}[t]
 \centering
 \includegraphics[scale=0.33]{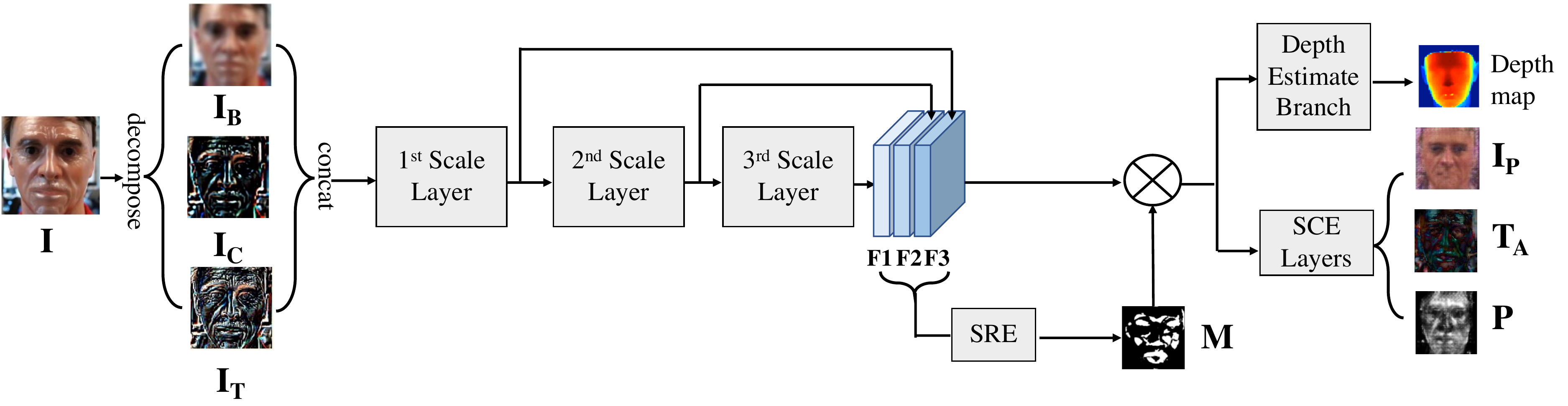}
 \caption{\footnotesize The ovearll SRENet architecture. We decompose the input image $\mathbf{I}$ into three elements (\textit{e.g.}, $\mathbf{I_{B}}$, $\mathbf{I_{C}}$ and $\mathbf{I_{T}}$), which represent the image information at different frequency levels. The multi-scale feature extractor takes the concatenation of these three elements to generate multi-scale features (\textit{e.g.}, $\mathbf{F_{1}}$, $\mathbf{F_{2}}$, and $\mathbf{F_{3}}$). Such multi-scale features are fed to the depth estimate branch for estimating the face depth, and SCE layers for estimating inpanting trace ($\mathbf{I_{p}}$), region trace ($\mathbf{P}$), and additive trace ($\mathbf{T_{A}}$). More importantly, SRE produces $\mathbf{M}$ to help pinpoint the spoof region. In the work~\cite{liu2020physics}, such SCE layers are three different branches, and three traces are used to synthesize the live counterpart of the input image via the adversarial training.}
 \label{fig_sreNet_architecture}
\end{figure}

%% file: figures_supp/siw_m_protocol_1.tex
\begin{table}[t!]
\small
\centering	
\begin{tabular}{ccc}
	\scalebox{0.6}{
	\begin{tabular}{cllll}
        \hline Protocol & Method & APCER (\%) & BPCER (\%) & ACER (\%) \\ \hline
        \multirow{3}{*}{1} 
               & PsySTD.~\cite{liu2020physics} & $0.0$ & $0.8$ & $0.4$ \\
               & PatchNet~\cite{wang2022patchnet} & $\mathbf{0.0}$ & $\mathbf{0.0}$ & $\mathbf{0.0}$ \\
               & Ours & $0.2$ & $0.6$ & $0.4$ \\ \hline
        \multirow{3}{*}{2} 
               & PsySTD.~\cite{liu2020physics} & $1.2$ & $1.3$ & $1.3$ \\
               & PatchNet~\cite{wang2022patchnet} & $\mathbf{1.1}$ & $1.2$ & $1.2$ \\
               & Ours & $1.4$ & $\mathbf{0.8}$ & $\mathbf{1.1}$ \\ \hline
        \multirow{3}{*}{3} 
               & PsySTD.~\cite{liu2020physics} & $1.7\pm1.4$ & $2.2\pm3.5$ & $1.9\pm2.3$ \\
               & PatchNet~\cite{wang2022patchnet} & $1.8\pm1.5$ & $\mathbf{0.6\pm1.2}$ & $\mathbf{1.2\pm1.3}$ \\
               & Ours & $\mathbf{1.6\pm1.6}$ & $1.2\pm1.4$ & $1.4\pm1.5$ \\ \hline
        \multirow{3}{*}{4} 
               & PsySTD.~\cite{liu2020physics} & $2.3\pm3.6$ & $4.2\pm5.4$ & $3.6\pm4.2$ \\
               & PatchNet~\cite{wang2022patchnet} & $2.5\pm3.8$ & $\mathbf{3.3\pm3.7}$ & $\mathbf{2.9\pm3.0}$ \\
               & Ours & $\mathbf{2.2\pm 1.9}$ & $3.8\pm 4.1$ & $3.0\pm 3.0$ \\ \hline
        \vspace{-2mm} \\
               \multicolumn{5}{c}{\large{(a)}}
    \end{tabular}}
    
    & & 
	\scalebox{0.7}{
	\begin{tabular}{cllll}
        \hline Protocol & Method & APCER (\%) & BPCER (\%) & ACER (\%) \\ \hline
        \multirow{4}{*}{1}
            & PsySTD.~\cite{liu2020physics} & $0.0$ & $0.0$ & $0.0$ \\
            & PatchNet~\cite{wang2022patchnet} & $0.0$ & $0.0$ & $0.0$ \\
            & Ours & $0.0$ & $0.0$ & $0.0$ \\ \hline
        \multirow{4}{*}{2}
            & PsySTD.~\cite{liu2020physics} & $0.0\pm0.0$ & $0.0\pm0.0$ & $0.0\pm0.0$ \\
            & PatchNet~\cite{wang2022patchnet} & $0.0\pm0.0$ & $0.0\pm0.0$ & $0.0\pm0.0$ \\
            & Ours & $0.0\pm0.0$ & $0.0\pm0.0$ & $0.0\pm0.0$ \\ \hline
        \multirow{4}{*}{3} 
            & PsySTD.~\cite{liu2020physics} & $13.1\pm9.4$ & $1.6\pm0.6$ & $7.4\pm4.3$ \\
            & PatchNet~\cite{wang2022patchnet} & $\mathbf{3.1\pm1.1}$ & $\mathbf{1.8\pm0.8}$ & $\mathbf{2.5\pm0.5}$ \\
            & Ours & $6.3\pm 1.3$ & $2.9\pm 0.4$ & $4.6\pm 0.9$ \\ \hline
        \vspace{-2mm} \\
            \multicolumn{5}{c}{\large{(b)}}
    \end{tabular}
        }\\
\end{tabular}
\vspace{1mm}
\caption{\footnotesize The baseline (SRENet) performance on (a) Oulu-NPU and (b) SiW datasets.}
\label{tab_oulu_siw}
\end{table}

%% file: figures_supp/siw_m_protocol_2.tex
\begin{table*}[t!]
\small
\centering
\begin{subtable}{1\linewidth}
    \resizebox{1\textwidth}{!}{
        \begin{tabular}{cc ccc cccccc cccc ccccc c}
	\multicolumn{18}{c}{} \\ \hline
	\multirow{2}{*}{Metric} && \multicolumn{4}{c}{Covering} && \multicolumn{3}{c}{Makeup}  && \multicolumn{5}{c}{3D Attack} && \multicolumn{2}{c}{2D Attack} && \multirow{2}{*}{Overall}\\
	\cline{3-6} \cline{8-10} \cline{12-16} \cline{18-19}
 	&
        & Fun. & Eye & Mou. & Pap. &
        & Ob. & Im. & Cos. &
        & Imp. & Sil. & Tra. & Pap. & Man. &
        & Rep. & Print &
        \\ \hline
        ACER(\%)&
        &$1.1$&$1.1$&$0.2$&$1.1$&
        &$0.0$&$3.6$&$2.7$&
        &$0.0$&$5.4$&$0.0$&$0.6$&$0.0$&
        &$1.9$&$1.5$& 
        &$2.6$ \\ \hline
        \begin{tabular}{c}
             TDR@  \\
             FDR=$1.0$(\%) 
        \end{tabular}&
        &$31.2$&$47.8$&$100.0$&$44.8$&
        &$100.0$&$80.0$&$87.5$&
        &$100.0$&$34.3$&$100.0$&$100.0$&$100.0$&
        &$97.4$&$98.2$&
        &$89.4$\\ \hline
        \end{tabular}
        \label{tab:siwm_p1}
    }
    \caption{\footnotesize{Protocol I: Unknown Spoof Attack Detection.} \vspace{-4mm}}
\end{subtable}

\begin{subtable}{1\linewidth}
    \resizebox{1\textwidth}{!}{
        \begin{tabular}{cc ccc cccccc cccc ccccc c}
    	\multicolumn{18}{c}{} \\ \hline
    	\multirow{2}{*}{Metric} && \multicolumn{4}{c}{Covering} && \multicolumn{3}{c}{Makeup}  && \multicolumn{5}{c}{3D Attack} && \multicolumn{2}{c}{2D Attack} && \multirow{2}{*}{Average}\\
    	\cline{3-6} \cline{8-10} \cline{12-16} \cline{18-19}
     	&
            & Fun. & Eye & Mou. & Pap. &
            & Ob. & Im. & Cos. &
            & Imp. & Sil. & Tra. & Pap. & Man. &
            & Rep. & Print &
            \\ \hline
        APCER(\%)&
        &$26.1$&$5.4$&$2.3$&$6.5$&
        & $2.7$&$6.1$&$8.0$&
        & $8.8$&$10.0$&$0.0$&$1.1$&$8.0$&
        & $19.9$&$2.7$& 
        &$7.7 \pm 7.0$ \\ \hline
        BPCER(\%)&
        &$33.0$&$0.0$& $0.0$& $17.3$&
        &$0.0$& $42.9$& $13.7$&
        &$7.1$& $8.5$& $0.0$& $0.0$& $0.0$&
        &$16.0$& $16.5$& 
        &$11.1 \pm 12.9$ \\ \hline
        ACER(\%)&
        &$29.5$&$2.7$&$1.1$&$11.9$&
        &$1.3$&$24.5$&$10.9$&
        &$8.0$&$9.2$&$0.0$&$0.6$&$4.0$&
        &$17.9$&$9.6$& 
        &$9.4 \pm 8.8$ \\ \hline
        \begin{tabular}{c}
             TDR@  \\
             FDR=$1.0$(\%) 
        \end{tabular}&
        &$8.9$&$37.0$&$88.4$&$4.0$&
        &$98.3$&$23.8$&$39.2$&
        &$61.4$&$47.4$&$100$&$100$&$66.6$&
        &$39.3$&$78.9$&
        &$56.7 \pm 32.0$\\ \hline
        \end{tabular}
        \label{tab:siwm_p2}
    }
    \caption{\footnotesize{Protocol II: Unknown Spoof Attack Detection.}\vspace{-4mm}}
\end{subtable}

\begin{subtable}{1\linewidth}
    \centering
    \resizebox{0.65\textwidth}{!}{
        \begin{tabular}{cc cccccccc ccc}
    	\multicolumn{10}{c}{} \\ \hline
    	\multicolumn{2}{c}{Metric} & \begin{tabular}{c}Source  \\Domain \end{tabular} & Spoof && Race && Age && Illum. & \multicolumn{3}{c}{Average}\\ \hline
        APCER(\%)
        &&$2.8$&$12.5$&&$18.9$&&$15.4$&&$7.7$
        &$11.5 \pm 5.7$& \\ \hline
        BPCER(\%)
        &&$1.5$&$17.4$&&$11.1$&&$0.0$&&$0.0$
        &$6.0 \pm 7.0$& \\ \hline
        ACER(\%)
        &&$2.2$&$14.9$&&$15.0$&&$7.7$&&$3.8$
        &$8.7 \pm 5.4$& \\ \hline
        \begin{tabular}{c}
             TDR@  \\
             FDR=$1.0$(\%) 
        \end{tabular}
        &&$86.2$&$28.3$&&$44.4$&&$55.6$&&$66.7$
        &$56.2 \pm 19.6$& \\ \hline
        \end{tabular}
        \label{tab:siwm_p3}
    }
    \caption{\footnotesize{Protocol III: Cross Domain Spoof Detection.}\vspace{-4mm}}
\end{subtable}
\vspace{1mm}
\caption{\footnotesize{The baseline (SRENet) performance on three protocols in SiW-Mv$2$ dataset.}}
\label{tab_siwm_overall}
\end{table*}